%% file: main.tex
\documentclass[letterpaper]{article} 
\usepackage[preprint]{aaai2027}  
\usepackage[hyphens]{url}  
\usepackage{graphicx}
\usepackage{natbib}  
\usepackage{caption} 
\usepackage{float}
\usepackage{booktabs}
\usepackage{amsmath,amssymb}
\usepackage{multirow}
\usepackage{xcolor}
\definecolor{pixcol}{HTML}{2F6E60} 
\definecolor{strcol}{HTML}{A34038} 
\definecolor{agtcol}{HTML}{34507E} 
\definecolor{envcol}{HTML}{5E6472} 
\definecolor{altcol}{HTML}{7A6BA6} 
\usepackage{tikz}
\usetikzlibrary{arrows.meta,positioning,fit,backgrounds,calc,decorations.pathreplacing,shadows.blur}
\usepackage{pgfplots}
\pgfplotsset{compat=1.18}
\usepgfplotslibrary{groupplots}
\usetikzlibrary{patterns}
\pgfplotsset{
  npaxis/.style={
    axis on top,
    tick align=outside,tick pos=left,
    axis line style={draw=black!45},
    every axis plot/.append style={line width=0.8pt},
    label style={font=\footnotesize},
    tick label style={font=\scriptsize},
    title style={font=\footnotesize\bfseries},
    ymajorgrids=true,grid style={dashed,draw=black!18},
    legend style={at={(0.5,1.02)},anchor=south,
                  draw=black!30,rounded corners=1pt,fill=white,
                  font=\scriptsize,inner sep=2.5pt,
                  /tikz/every even column/.append style={column sep=5pt}},
  },
  npbarlegend/.style={
    legend image code/.code={\draw[#1] (0cm,-0.07cm) rectangle (0.22cm,0.13cm);}},
}
\newcommand{\imgicon}{\tikz[baseline=-0.35ex]{%
  \draw[line width=0.35pt,pixcol,fill=pixcol!16,rounded corners=0.3pt]
        (0,0) rectangle (1.7ex,1.35ex);
  \fill[pixcol] (0.42ex,0.95ex) circle (0.16ex);
  \draw[pixcol,line width=0.35pt,line join=round]
        (0.2ex,0.25ex)--(0.65ex,0.82ex)--(1.02ex,0.5ex)--(1.5ex,1.0ex);}}
\newcommand{\codeicon}{\textcolor{strcol}{\ttfamily </>}}

\graphicspath{{figures/}}

\AtBeginDocument{\raggedbottom}

\newcommand{\PFR}{\textsc{pfr}}
\newcommand{\SFR}{\textsc{sfr}}
\newcommand{\PFG}{\textsc{pfg}}
\newcommand{\VOR}{\textsc{vor}}
\newcommand{\PPD}{\textsc{ppd}}
\newcommand{\SRM}{\textsc{srm}}
\newcommand{\NVR}{\textsc{nvr}}
\newcommand{\Accimg}{\mathrm{Acc}_{\mathrm{img}}}
\newcommand{\Acczero}{\mathrm{Acc}_{0}}
\newcommand{\Acccrop}{\mathrm{Acc}_{\mathrm{crop}}}
\newcommand{\task}{\emph{visual state reliance}}
\newcommand{\Task}{\emph{Visual state reliance}}

\title{Do GUI Agents Believe Their Eyes? Diagnosing State-Belief Reliance on Pixels versus Structure}
\author{
    Gui\kern0.05pt jia Zhang\textsuperscript{\rm 1,2}\qquad
    Yuxun Chen\textsuperscript{\rm 1}\qquad
    Yuheng Qi\textsuperscript{\rm 1}\qquad
    Harry Yang\textsuperscript{\rm 2}
}
\affiliations{
    \textsuperscript{\rm 1}Shenzhen University\\
    \textsuperscript{\rm 2}The Hong Kong University of Science and Technology
}

\begin{document}
\maketitle

\begin{abstract}
\input{sections/00_abstract}
\end{abstract}

\input{sections/01_intro}
\input{sections/02_related}
\input{sections/03_formulation}
\input{sections/04_benchmark}
\input{sections/05_metrics}
\input{sections/06_experiments}
\input{sections/07_discussion}
\input{sections/08_conclusion}

\bibliography{refs}

\clearpage
\appendix
\setcounter{secnumdepth}{2}
\section*{Technical Appendix}
\input{sections/09_appendix}

\end{document}

%% file: sections/00_abstract.tex
Multimodal GUI agents read an interface through two redundant channels: the
rendered \emph{pixels} of a screenshot and a serialized \emph{structure}
such as a document object model or accessibility tree. Before acting, an agent forms a belief
about the current interface state, but existing benchmarks score task
success, element grounding, or attack resistance and do not ask whether that
belief is drawn from the pixels. We formalize \task{}, the attribution of a
state belief to pixels, structure, or priors, and measure it with paired
single-channel interventions over $735$ probes spanning real web, mobile,
and desktop interfaces, of which $225$ are zero-edit divergences mined from
live production websites, all scored by deterministic forced choice with no
model judge. Our central metric is the Perception-Fusion Gap \PFG{}, the
fraction of probes a model perceives correctly yet resolves toward structure
under conflict; a stricter variant that re-verifies perception on a tight
crop of the target region leaves the gap intact. Across models from four
vendors, textual state beliefs defer to structure while image-only accuracy
stays near ceiling, and on unedited stale snapshots from live pages the same
models follow the outdated structure on up to $0.88$ of probes. A white-box
ablation traces the textual effect to a single copied structural value, and
gradient attribution shows the visual evidence is processed yet overridden.
In live multi-step environments one mis-sourced belief at the first step
compounds into task failure with a self-recovery rate of at most $0.03$.
Comparing four mitigations on identical probes, prompt-level cues fail at
the action level, certificate checks buy safety with refusals, and a
training-free consistency gate is alone in reducing both hijack and task
error. \Task{} thus gives a measurable diagnostic of whether agent state
beliefs are visually grounded.

%% file: sections/01_intro.tex
\section{Introduction}
\label{sec:intro}

\begin{figure}[t!]
\centering
\begin{tikzpicture}[
  font=\footnotesize,
  >={Stealth[length=2.4mm]},
  chanbox/.style={rounded corners=4pt,draw=none,align=left,inner sep=4pt,
               blur shadow={shadow blur steps=5,shadow xshift=0pt,
               shadow yshift=-1pt,shadow opacity=18}},
  tabY/.style={rounded corners=3pt,fill=envcol!15,draw=none,
               text=envcol!95!black,font=\scriptsize\bfseries,inner sep=2.5pt},
  neqchip/.style={rounded corners=3pt,fill=strcol!14,draw=none,
               text=strcol,font=\scriptsize\bfseries,inner sep=2.5pt},
  agentbox/.style={rounded corners=5pt,draw=none,
               fill=agtcol!13,align=center,text width=6.2cm,minimum height=0.8cm,
               font=\bfseries,blur shadow={shadow blur steps=5,shadow xshift=0pt,
               shadow yshift=-1pt,shadow opacity=22}},
  belief/.style={rounded corners=4pt,draw=none,align=center,inner sep=3pt,
               text width=2.95cm,minimum height=0.8cm,
               blur shadow={shadow blur steps=5,shadow xshift=0pt,
               shadow yshift=-1pt,shadow opacity=20}},
  arlbl/.style={font=\scriptsize\itshape,text=black!55,fill=white,inner sep=1pt},
]
\newcommand{\chev}[1]{%
  \draw[line width=1.5pt,agtcol,line cap=round,line join=round]
    ($(#1)+(-0.17,0.15)$) -- ($(#1)+(0,0.01)$) -- ($(#1)+(0.17,0.15)$);
  \draw[line width=1.5pt,agtcol,line cap=round,line join=round]
    ($(#1)+(-0.17,-0.04)$) -- ($(#1)+(0,-0.18)$) -- ($(#1)+(0.17,-0.04)$);}
\node[chanbox,fill=pixcol!11,text width=6.1cm] (pix){%
  \makebox[\linewidth][c]{\bfseries\color{pixcol}\imgicon\ Pixels $P$: screenshot}\\[2pt]
  \makebox[\linewidth][c]{\includegraphics[width=5.55cm]{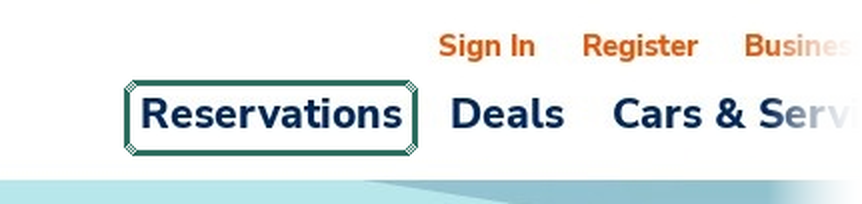}}\\[1.5pt]
  \makebox[\linewidth][c]{\tikz{\node[neqchip]{conflict: $P \neq S$};}}};
\node[chanbox,fill=strcol!9,text width=6.1cm,below=1.6mm of pix] (str){%
  \makebox[\linewidth][c]{\bfseries\color{strcol}\codeicon\ Structure $S$: DOM (edited)}\\[2pt]
  {\ttfamily\footnotesize <a role="button">}\\[1pt]
  {\ttfamily\footnotesize \ \ <text>\,%
     \colorbox{strcol!14}{\textcolor{strcol}{Budget Truck}}</text>}\\[1pt]
  {\ttfamily\footnotesize </a>}};
\begin{scope}[on background layer]
\node[rounded corners=6pt,draw=none,fill=envcol!7,inner sep=4pt,
      fit=(pix)(str)] (env){};
\end{scope}
\node[tabY,anchor=south] at ([yshift=-2pt]env.north)
      {Environment state: \imgicon\,visual $+$ \codeicon\,textual};

\node[agentbox,below=4.5mm of env] (ag) {Multimodal GUI agent\quad$\hat y=g(P,S)$};
\coordinate (envag) at ($(env.south)!0.5!(ag.north)$);
\chev{envag}

\coordinate (by) at ([yshift=-5mm]ag.south);
\node[belief,fill=pixcol!11,anchor=north west]
      (bp) at (env.south west |- by){%
  {\bfseries\color{pixcol}Pixels-only read}\\[2pt]
  \textsf{Reservations}~\textcolor{pixcol}{\checkmark}\\[1pt]
  {\itshape perception correct}};
\node[belief,fill=strcol!9,anchor=north east]
      (bf) at (env.south east |- by){%
  {\bfseries\color{strcol}Fused belief}\\[2pt]
  \textcolor{strcol}{\textsf{Budget Truck}}~\textcolor{strcol}{\boldmath$\times$}\\[1pt]
  {\itshape follows structure}};
\draw[-{Stealth[length=2.6mm]},line width=1.2pt,pixcol,dashed]
      ([xshift=-3mm]ag.south) to[out=-130,in=90] (bp.north);
\draw[-{Stealth[length=2.6mm]},line width=1.2pt,strcol]
      ([xshift=3mm]ag.south) to[out=-50,in=90] (bf.north);
\node[arlbl] at ([xshift=-2.15cm,yshift=-0.32cm]ag.south) {isolated};
\node[arlbl] at ([xshift= 2.15cm,yshift=-0.32cm]ag.south) {fused};

\draw[decorate,decoration={brace,amplitude=4pt,mirror},thick,black!55]
  ([yshift=-2pt]bp.south west) -- ([yshift=-2pt]bf.south east)
  node[midway,below=4pt,align=center,font=\scriptsize]
  {\textbf{\PFG}: perceive correctly yet defer};
\end{tikzpicture}
\caption{The core scenario, on a real probe from
Multimodal-Mind2Web~\citep{deng2023mind2web}. The same interface state reaches the
agent through two channels: the rendered \emph{pixels} $P$, a screenshot whose
highlighted control reads \textsf{Reservations}, and the serialized \emph{structure}
$S$, a DOM node we edit to read \textsf{Budget Truck}. Read from the pixels alone the
agent answers \textsf{Reservations} correctly, yet its \emph{fused} belief follows
the edited structure. We report the fraction of such perceive-correctly-yet-defer
probes as the Perception--Fusion Gap \PFG, our central metric.}
\label{fig:teaser}
\end{figure}

Multimodal GUI agents now operate computers, browsers, and phones by
consuming two redundant views of the same interface: the rendered
\emph{pixels} of a screenshot and a serialized \emph{structure}, the textual
description a platform exposes alongside the screen, such as the browser's
document object model or an accessibility
tree~\citep{deng2023mind2web,koh2024visualwebarena}. Before an agent can plan
or act, it must form a \emph{belief about the current state of the
interface}: whether a toggle is on, which tab is selected, or what text a
field contains. This belief precedes every action, and a wrong belief
propagates into the decisions that follow even when the planner is otherwise
competent. Existing benchmarks score the \emph{end} of that loop, namely task
completion, element grounding, or robustness to injected
attacks~\citep{koh2024visualwebarena,cheng2024seeclick,li2025vpibench}, and so
leave the origin of the belief unmeasured.

The two channels are usually consistent, yet they diverge often enough to
matter: accessibility trees go stale, content renders after the structure is
captured, and serialized labels disagree with displayed text. In
CLAY~\citep{li2022clay}, a human re-annotation of the RICO corpus of Android
screens, $10.6\%$ of structure nodes have \emph{no valid visual
representation}, and such a ghost node appears on $37.4\%$ of screens. Mining
$38$ live production websites with no edit of any kind, we collected $225$
elements whose serialized value disagrees with the rendered one, among them
ticker, clock, and counter values whose captured structure was outdated
within seconds. A model that reports the correct state by copying a
structural string it never visually verified matches the screen only as long
as the structure happens to agree, and it fails without warning once
structure and reality diverge. Because uncurated divergences alone cannot
reveal which channel a belief follows, we pair them with controlled
single-channel edits that isolate the mechanism.

We therefore ask a question these benchmarks cannot answer: \emph{is a
multimodal agent's belief about the visible state actually sourced from the
pixels?} This differs from whether the model \emph{can} perceive the state,
since a model may read a screenshot perfectly in isolation and still report
a conflicting structural value once one is present. Separating perception
from fusion requires paired interventions that change one channel at a time,
together with a metric for the gap between them. We operationalize this as
\task{}, show that the substitution occurs even where perception is
demonstrably correct, and trace its errors into actions and multi-step
trajectories. The effect is not uniform: it is strongest for \emph{textual}
state, where a serialized string overrides a correctly read screen, whereas
non-text identity such as icons and widget types stays largely pixel-bound
in capable models.

\paragraph{Findings.}
Four open-weight VLMs run locally form the reproducible core, and three
OpenAI models corroborate. On textual web state every model defers to a
conflicting structural value while its image-only accuracy stays near
ceiling, so the Perception-Fusion Gap is positive throughout, and a stricter
variant that re-verifies perception on a tight crop of the target region
leaves the gap essentially unchanged. The behavior is not an artifact of our
edits: on zero-edit divergences mined from live websites, the same models
follow an outdated structural snapshot on up to $0.88$ of stale probes, and
a severity analysis places our synthetic conflicts inside the distribution
of natural ones. Ablating exactly the conflicting value flips the belief
back to the pixels far more often than an equal-sized random ablation, so
the failure is one of fusion rather than perception. The mis-sourced belief
then compounds: in live multi-step episodes a single first-step conflict
drives the task to failure with a self-recovery rate of at most $0.03$, and
of four mitigations compared on identical action probes only a training-free
consistency gate that re-grounds on a detected conflict reduces hijack and
task error together.

\paragraph{Contributions.}
\begin{itemize}\setlength{\itemsep}{2pt}
  \item We formalize \task{} and define the Perception-Fusion Gap \PFG{},
        which conditions on correct perception to separate what a model
        \emph{sees} from what it \emph{believes} under a conflicting
        channel, with a crop-verified strict variant that guards the
        construct.
  \item We build a $735$-probe benchmark over real web, mobile, and desktop
        data whose conflicts are deterministic rule-based edits or left
        untouched, including $225$ zero-edit divergences mined from $38$
        live websites and $250$ stale-node candidates from human-annotated
        mobile corpora, validated by double annotation at $\kappa{=}0.86$
        and a severity analysis.
  \item We show that textual state beliefs defer to structure across models
        from four vendors despite correct pixel perception, trace the effect
        to a single copied structural value by white-box ablation, and show
        by gradient attribution that visual evidence is processed yet
        overridden.
  \item We quantify downstream harm and defenses: one mis-sourced belief
        compounds into near-certain failure of live multi-step episodes, and
        a four-way mitigation comparison on identical probes shows only a
        training-free consistency gate reduces hijack and task error
        together.
\end{itemize}

%% file: sections/02_related.tex
\section{Related Work}
\label{sec:related}

\paragraph{Agent benchmarks score the end of the act loop.}
Mind2Web~\citep{deng2023mind2web},
VisualWebArena~\citep{koh2024visualwebarena},
OSWorld~\citep{xie2024osworld}, AndroidWorld~\citep{rawles2024androidworld},
and GUI-World~\citep{chen2024guiworld} report task success;
SeeClick/ScreenSpot~\citep{cheng2024seeclick} and
ScreenSpot-Pro~\citep{li2025screenspotpro} report element localization. Such
scores can stay high when redundant structure supplies the answer, never
revealing which channel the model trusted. We reuse these interfaces as
substrate but change one channel at a time, so the measured quantity is the
source of the belief, not execution skill.

\paragraph{GUI grounding measures accuracy, not provenance.}
Because binding a plan to a concrete element is the bottleneck rather than
planning~\citep{zheng2024seeact}, screenshot-first agents de-emphasize long
DOM input~\citep{he2024webvoyager}, driving a wave of vision-first
grounding: universal visual grounding~\citep{gou2025uground}, cross-platform
action models~\citep{wu2025osatlas,hong2024cogagent}, native GUI
agents~\citep{qin2025uitars,xu2025aguvis,lin2025showui}, pixel-to-structure
parsers~\citep{lu2024omniparser}, and attention-guided grounding at high
resolution~\citep{li2025screenspotpro,wu2025guiactor,yuan2025segui};
conversely, DOM snapshots are matched to screenshots~\citep{d2snap2025},
pixels re-parsed into accessibility trees~\citep{screen2ax2025}, and
accessibility input compressed and fused~\citep{takeshita2026a11y}. None of
this can say \emph{which} channel a belief used when a correct screen and a
conflicting structure coexist; ``pure vision'' is a training claim, and the
same agents still receive auxiliary structure at
inference~\citep{jia2025osworldmcp}.

\paragraph{Internal-prior studies pit pixels against language.}
Models drift from images toward language priors~\citep{morethinking2025},
step-level beliefs may not be visually grounded~\citep{steplevel2026}, and
similar shortcuts appear in driving~\citep{drivingvla2026} and agent
memory~\citep{memeye2026}. These lines pit pixels against the model's
\emph{internal} prior and read aggregate accuracy; we pit pixels against an
\emph{external} structured channel that diverges in deployment, and our
paired intervention conditions on correct perception of the very probe it
tests.

\paragraph{Safety, injection, and action-level defenses.}
A third thread stresses agents with adversarial content and scores harmful
behavior~\citep{li2025vpibench,eva2025,envinjection2025,osharm2025}, while
action-level defenses allow a tool call only when cross-modal certificates
corroborate each predicate~\citep{eca2026}. These define an action and an
adversary, and they \emph{presuppose} the upstream rate at which a
pixel-unsupported value enters the belief, which is exactly the gap we
measure; our diagnostic also tells the two failure modes apart, since a
pixels-only agent is blind to structure-borne threats that never render
while a structure-reading agent inherits whatever the channel asserts.
Closest in spirit, the Visual Confused Deputy~\citep{vcd2026} perturbs the
rendered screen to hijack a computer-use agent; our manipulation is the
inverse and benign, holding the pixels correct and conflicting the
structured channel.

%% file: sections/03_formulation.tex
\section{Problem Formulation}
\label{sec:formulation}

\paragraph{State probes.}
A \emph{state probe} is a tuple
$(\,\text{id},\, I,\, S,\, r,\, v,\, a^\star)$, where $I$ is a real
screenshot, $S$ is a serialized structure such as a DOM, accessibility tree,
or view hierarchy, $r$ is the target region's bounding box, $v$ is a single
discrete state variable with finite value set $V_v$, and
$a^\star\in V_v$ is the \emph{gold} value. In the running example of
Fig.~\ref{fig:teaser}, $v$ is the text of a highlighted control, the pixels
render $a^\star{=}$\textsf{Reservations}, and the structure is edited to
$b{=}$\textsf{Budget Truck}. We fix a deliberate stance: \textbf{gold is
always defined by the pixels}, the value a human sees on screen, and we ask
which channel the model's belief \emph{follows}.

\paragraph{Channels.}
A belief can be sourced from the \emph{pixels} $P$, from the \emph{structure}
$S$, or from a \emph{prior}, the model's default when neither channel is
informative. OCR and captioning are treated as sub-cases of $S$ in ablations.

\paragraph{Paired channel interventions.}
For each probe we derive the conditions of Table~\ref{tab:conditions}, which
hold the content fixed and switch exactly one channel. The headline
\emph{structure-swap} contrast pairs a real screenshot ($P=a^\star$) with an
edited structure ($S=b\neq a^\star$), so following structure means answering
$b$; the remaining conditions provide the baseline, single-channel readings,
artifact and prior controls, and the unedited natural conflict.

\begin{table}[t]
\centering\footnotesize
\begin{tabular}{@{}llll@{}}
\toprule
Condition & $P$ & $S$ & Purpose \\
\midrule
agreement        & $a^\star$       & $a^\star$ & baseline \\
structure-swap   & $a^\star$       & $b$       & \textbf{substitution} \\
natural-conflict & $a^\star$       & present   & natural conflict \\
pixel-swap       & $b$             & $a^\star$ & pixel sensitivity \\
re-render        & $a^\star{\to}a^\star$ & $a^\star$ & artifact control \\
pixels-only      & $a^\star$       & $\varnothing$ & perception \\
structure-only   & $\varnothing$   & $b$       & structure read \\
no-evidence      & occluded        & $\varnothing$ & prior \\
\bottomrule
\end{tabular}
\caption{Paired interventions per probe. Only one channel changes at a time,
so any difference between conditions is attributable to that channel.}
\label{tab:conditions}
\end{table}

\paragraph{Output schema.}
The model returns forced-choice JSON: an \texttt{answer} $\in V_v$, a
\texttt{confidence}, and a self-reported evidence channel. Only the
deterministic match of \texttt{answer} against $a^\star$ or $b$ enters the
main metrics; no LLM judge is used.

\paragraph{State families.}
Probes span five conflict families across three platforms so that
``following pixels'' cannot collapse into string matching: \emph{web-text}
covers textual values on web, \emph{mobile-widget} covers non-text control
identity on mobile such as icon, image, and switch type,
\emph{mobile-stale} covers stale or ghost accessibility-tree references on
mobile, \emph{desktop-graphic} covers icon-versus-label identity on
desktop, and \emph{web-natural} covers unedited divergences mined from
live websites. The two non-text families are decisive, since a model that
follows structure there is not merely reading text.

\paragraph{Interventional interpretation.}
Writing the belief as a deterministic $\hat{y}=g(P,S)$, the paired design
changes only one argument, so answering $a^\star$ under agreement but $b$
under structure-swap is interventional evidence that $S$ drives the belief,
with the pairing controlling for task, layout, and model identity.

%% file: sections/04_benchmark.tex
\section{Benchmark Construction}
\label{sec:benchmark}

\paragraph{Sources, real data only.}
Web probes are built from Multimodal-Mind2Web~\citep{deng2023mind2web}, which
pairs real screenshots with the full DOM. Mobile probes are built from
RICO~\citep{deka2017rico} view hierarchies augmented with
CLAY~\citep{li2022clay} human annotations, which flag structure nodes that
annotators judge invalid or background, that is, stale or ghost entries in
the accessibility tree. Desktop probes are built from
ScreenSpot-Pro~\citep{li2025screenspotpro}, a high-resolution suite of
professional applications. A further family is mined directly from live
production websites as described below. No screenshot, label, or structure
is generated by a language model; every conflict is produced by a
deterministic rule-based edit or is naturally present in the data.

\paragraph{Edited-conflict families.}
In the \emph{web-text} family we locate a bounding-box-verified text node
and build a structure-swap conflict by replacing the structural string with
a plausible on-page distractor, leaving the screenshot untouched. The
\emph{mobile-widget} family takes a non-text widget such as an icon, image,
or switch, whose CLAY type label is the pixel gold while the structure
asserts a text label. For \emph{desktop-graphic} we use the ScreenSpot-Pro
label distinguishing an icon from a text label as pixel gold and assert the
opposite type in the structure. Mobile visible controls serve as a
perception baseline.

\paragraph{Natural-conflict families, zero edits.}
Two families carry divergences we do not create. The \emph{web-natural}
family is mined from $38$ live production websites by a crawler that flags
three kinds of divergence between the serialized DOM and the rendered page:
an \texttt{aria-label} whose text materially disagrees with the visible
label, a \texttt{title} attribute that materially disagrees with the visible
text, and \emph{stale} snapshots, where the DOM is captured and the page
then updates a ticker, clock, or counter before the screenshot, exactly the
capture race agent frameworks face. Materially means the normalized strings
share no containment relation, excluding benign descriptive labels; each of
the $225$ probes archives the full page and DOM. The \emph{mobile-stale}
family takes CLAY-flagged invalid nodes whose structure lists an element the
screen may not render, expanded to $250$ candidates over $99$ screens; since
CLAY flags are noisy, scoring uses only the subset confirmed absent by
unanimous human audit.

\paragraph{Severity of the edits.}
A synthetic distractor could be unlike anything deployment produces, which
would make the measured deference unrepresentative. We therefore compare
the edited pairs against the $225$ natural divergence pairs under five
string severity measures, among them normalized edit distance, token
Jaccard distance, and length difference. The synthetic pairs fall inside
the natural $5$th-to-$95$th percentile band at a coverage of $0.93$ to
$1.00$ per measure, and the synthetic medians sit at or slightly above the
natural ones: the edits operate in the severity regime live pages exhibit,
and where they differ they are the more blatant, not the subtler, case.

\paragraph{Composition.}
The benchmark has $735$ probes: $120$ web-text, $50$ mobile-widget, $60$
desktop-graphic, $30$ mobile visible controls, $250$ mobile-stale
candidates, and $225$ web-natural. The edited-conflict headline uses the
three edited families, $230$ probes; the natural families are scored
separately since their gold requires visibility or provenance audit.

\paragraph{Gold-label audit.}
Each probe is auto-constructed and then audited against the rendered pixels
by two independent annotators who see only the screenshot and never the
structure. Over all $735$ probes the annotators agree at $0.880$ raw
agreement and Cohen's $\kappa{=}0.86$. Validity, the share of audited probes
whose pixel judgment matches the proposed gold, is $0.94$ on web text,
$0.96$ on mobile widgets, $0.96$ to $0.97$ on web-natural, and $0.83$ to
$0.88$ on desktop-graphic, the weakest family; the Experiments section
therefore re-scores desktop on the clean subset both annotators confirm, and
the conclusions survive. Web-natural is likewise scored on its
audit-consensus subset of $182$ probes, since stale ticker values render
smallest on screen, and mobile-stale is scored only on its unanimously
confirmed-absent subset, because its proposed gold asserts absence. Every condition uses a neutral forced-choice prompt
with shuffled options, matched by string identity with no model judge.

%% file: sections/05_metrics.tex
\section{Metrics}
\label{sec:metrics}

Let $\hat{y}^c_i$ be the forced-choice answer of probe $i$ under condition
$c$, with pixel truth $a^\star_i$ and structural conflict value $b_i$. We
write the conditions by name, using $\text{pix}$ for pixels-only, $\text{agree}$
for the agreement baseline, $\text{swap}$ for the structure-swap conflict, and
$\text{pxsw}$ for pixel-swap.

\paragraph{Perception and channel following.}
$\Accimg=\frac1N\sum_i\mathbb{1}[\hat{y}^{\text{pix}}_i=a^\star_i]$ measures
whether the model \emph{can} see the state, and $\Acczero$, the analogous
accuracy under agreement, checks solvability. Under the structure-swap
conflict the pixel-following rate
$\PFR=\frac1N\sum_i\mathbb{1}[\hat{y}^{\text{swap}}_i=a^\star_i]$ and the
structure-following rate
$\SFR=\frac1N\sum_i\mathbb{1}[\hat{y}^{\text{swap}}_i=b_i]$ track which
channel the belief follows; \SFR{} is the primary substitution signal, and
the remainder $1-\PFR-\SFR$ is other or abstain.

\paragraph{Perception-Fusion Gap.}
Restricted to the perception-correct subset
$\mathcal{C}=\{i:\hat{y}^{\text{pix}}_i=a^\star_i\}$,
\begin{equation}
\PFG=\frac{1}{|\mathcal{C}|}\sum_{i\in\mathcal{C}}
      \mathbb{1}[\hat{y}^{\text{swap}}_i=b_i].
\end{equation}
\PFG{} is the fraction of probes the model \emph{sees correctly} yet, in the
fused setting, hands to structure. It separates a perception failure from a
fusion failure, the distinction this work is built on.

\paragraph{A strict variant guards the construct.}
Conditioning on the full-screenshot pixels-only read leaves one loophole: a
model might pass that check using page context while never resolving the
target region, in which case \PFG{} would overstate fusion failure. We
therefore add a condition showing only a tight crop of the target region
with no structure, and define $\PFG_{\text{strict}}$ by conditioning on
probes answered correctly from the crop alone; perception verified on the
isolated region cannot be an artifact of surrounding context. The two
versions agree closely on every open model, so the headline \PFG{} is not
inflated by this loophole.

\paragraph{Statistics.}
We report bootstrap and cluster-robust $95\%$ confidence intervals and test
model-pair differences in structure-following with the exact McNemar test
under Holm correction.

%% file: sections/06_experiments.tex
\section{Experiments}
\label{sec:experiments}

We proceed in five steps: establish the effect, verify the construct,
confirm it on natural unedited divergences, localize the cause with a
white-box ablation, and trace the belief into live actions, multi-step
episodes, and a controlled comparison of mitigations.

\paragraph{Setup.}
The reproducible core is four open-weight VLMs run locally from fixed
checkpoints: Qwen2.5-VL-7B-Instruct and Qwen3-VL-30B-A3B from Alibaba,
InternVL3-8B from OpenGVLab, and OS-Atlas-Base-7B from Shanghai AI Lab;
every headline claim holds on open weights alone. Three OpenAI models,
gpt-5.4, gpt-4o, and gpt-5.4-nano, queried in June 2026, enter only as
corroboration on the original $310$ probes, since such endpoints can drift.
All models share prompts, a deterministic parser, temperature-$0$ decoding
with one sample per probe and condition, and shuffled options. Web results
use the web-text family of $N{=}120$ unless noted.

\subsection{Textual state defers under correct perception}
Table~\ref{tab:main} reports the web results. Every image-reading model
perceives the target almost perfectly, with $\Accimg$ between $0.85$ and
$0.93$ and the agreement baseline near ceiling, yet under the structure-swap
conflict each shows a positive Perception-Fusion Gap, from $0.30$ for
Qwen2.5-VL-7B to $0.75$ for InternVL3-8B. Even the most pixel-faithful model
flips $0.30$ of the probes it perceives correctly. Since perception and
solvability stay near ceiling, the gap is a fusion failure, and it holds
across API and open weights with a model-dependent magnitude. Cluster-robust
$95\%$ intervals that resample by website keep \PFG{} strictly positive for
every model, with lower bounds from $0.17$ to $0.68$; Appendices~\ref{app:cluster} and~\ref{app:mcnemar}
give the full tables and paired tests.

\begin{table}[t]
\centering\small
\setlength{\tabcolsep}{3.3pt}
\begin{tabular}{@{}llccccc@{}}
\toprule
Model & Vendor & $\Accimg$ & \PFR & \SFR & \PFG & $\PFG_{\text{str}}$ \\
\midrule
gpt-5.4               & OpenAI    & 0.93 & 0.45 & \textbf{0.55} & \textbf{0.51} & -- \\
gpt-4o                & OpenAI    & 0.92 & 0.28 & \textbf{0.72} & \textbf{0.69} & -- \\
gpt-5.4-nano          & OpenAI    & 0.90 & 0.39 & \textbf{0.61} & \textbf{0.57} & -- \\
Qwen2.5-VL-7B         & Alibaba   & 0.86 & 0.59 & \textbf{0.41} & \textbf{0.30} & 0.34 \\
Qwen3-VL-30B          & Alibaba   & 0.85 & 0.23 & \textbf{0.77} & \textbf{0.71} & 0.77 \\
InternVL3-8B          & OpenGVLab & 0.89 & 0.21 & \textbf{0.79} & \textbf{0.75} & 0.75 \\
OS-Atlas-7B           & SH AI Lab & 0.92 & 0.49 & \textbf{0.51} & \textbf{0.45} & 0.44 \\
\midrule
structure-only        & OpenAI    & --   & 0.06 & 0.94 & -- & -- \\
\bottomrule
\end{tabular}
\caption{Web-text results, $N{=}120$. Every image-reading model perceives
near ceiling yet shows a positive \PFG{}; the crop-verified strict variant
$\PFG_{\text{str}}$ matches \PFG{} closely and is computed on the fixed
open checkpoints only, since the crop condition postdates the API runs and
rolling endpoints drift. The blind structure-only reader marks the
structure-following ceiling.}
\label{tab:main}
\end{table}

\paragraph{A strict perception check.}
Recomputing the gap on probes answered correctly from the tight crop alone
addresses the loophole of the Metrics section. Crop accuracy is $0.91$ to
$0.93$ on web text, higher than the full-screenshot read, and
$\PFG_{\text{strict}}$ stays within $0.05$ of \PFG{} on every open model,
rising for three of the four, so the strict check does not shrink the gap. Gradient attribution corroborates from inside the model:
among web-text conflicts where the belief follows structure, the attribution
mass on visual tokens is non-zero on every single probe and comparable in
magnitude to pixel-following cases, $0.028$ versus $0.034$ on Qwen2.5-VL-7B
with the same pattern on the other instrumented models, tabulated in
Appendix~\ref{app:strict}. The visual evidence is processed and then overridden at
fusion.

\subsection{Natural conflicts in the wild}
\label{sec:natural}
Everything above uses controlled edits, so we next ask whether deployment
produces divergences that trigger the same behavior, using the two zero-edit
families. Table~\ref{tab:natural} splits the web-natural family by kind on
its audit-consensus subset, and the pattern is sharply bimodal. When the
divergent structural value is descriptive metadata that paraphrases rather
than mirrors the visible text, models overwhelmingly report the rendered
value, with structure-following at most $0.08$. When it is a \emph{stale
snapshot} of the same field, a ticker or clock the structure captured
moments before the screenshot, the same models follow the outdated structure
on $0.38$ to $0.88$ of probes, with $95\%$ lower confidence bounds of $0.25$
to $0.75$ over $42$ probes from $11$ sites. The dangerous case is precisely
the one deployment produces on every dynamic page: the structure asserts a
plausible but outdated value of the very field the agent must read, and the
belief copies it.

\begin{table}[t]
\centering\small
\setlength{\tabcolsep}{4.5pt}
\begin{tabular}{@{}lcccc@{}}
\toprule
 & & \multicolumn{3}{c}{\SFR{} by divergence kind} \\
\cmidrule(l){3-5}
Model & $\Accimg$ & aria & title & stale \\
\midrule
Qwen2.5-VL-7B & 0.94 & 0.02 & 0.03 & \textbf{0.38} \\
Qwen3-VL-30B  & 0.96 & 0.02 & 0.01 & \textbf{0.76} \\
InternVL3-8B  & 0.95 & 0.05 & 0.02 & \textbf{0.88} \\
OS-Atlas-7B   & 0.96 & 0.02 & 0.08 & \textbf{0.45} \\
\bottomrule
\end{tabular}
\caption{Web-natural family, the $182$-probe audit-consensus subset of $225$
zero-edit divergences from $38$ live websites. Descriptive aria and title
divergences barely move the belief; stale snapshots of the queried field
hijack it at $0.38$--$0.88$.}
\label{tab:natural}
\end{table}

The mobile-stale family gives the natural test statistical teeth. Two human
auditors annotated all $250$ candidates, agreeing at $0.912$ raw agreement
and $\kappa{=}0.825$, and unanimously confirm $114$ elements truly absent
from the screen. On this audited-absent subset the models report the ghost
element visible at \SFR{} $0.99$, $0.94$, and $0.75$ for Qwen2.5-VL-7B,
InternVL3-8B, and OS-Atlas-7B, with lower confidence bounds of $0.95$,
$0.88$, and $0.66$. Conditioning as in \PFG{} on probes whose image-only
read already detects the absence, the fused belief still flips to visible on
$0.98$, $0.93$, and $0.63$ of them: the natural analogue of the headline
gap, measured with zero edits; full per-kind tables are in
Appendix~\ref{app:natural}. Qwen3-VL-30B is the instructive exception,
largely immune here at \SFR{} $0.19$ while being the \emph{most} hijacked
model on synthetic web text and among the most hijacked on the stale-web
kind at $0.76$; capability repairs visibility grounding but not text-value
fusion, so scale alone does not close the failure. With the severity
analysis of the Benchmark section, this closes the gap between motivation
and evidence: natural divergences are frequent, our edits are no harsher,
and the harshest natural kinds hijack beliefs at rates matching the
synthetic headline.

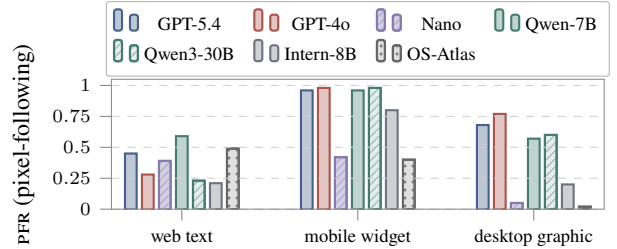
\begin{figure}[t]
\centering
\begin{tikzpicture}
\begin{axis}[npaxis,npbarlegend,
  ybar,bar width=4.4pt,
  width=8.1cm,height=3.3cm,
  ymin=0,ymax=1.05,enlarge x limits=0.2,
  symbolic x coords={web text,mobile widget,desktop graphic},
  xtick=data,ytick={0,0.25,0.5,0.75,1},
  ylabel={\PFR{} (pixel-following)},legend columns=4,
]
\addplot[draw=agtcol!90,fill=agtcol!32] coordinates {(web text,0.45)(mobile widget,0.96)(desktop graphic,0.68)};
\addplot[draw=strcol!90,fill=strcol!30] coordinates {(web text,0.28)(mobile widget,0.98)(desktop graphic,0.77)};
\addplot[draw=altcol!90,fill=altcol!32,
         postaction={pattern=north east lines,pattern color=altcol!60}]
  coordinates {(web text,0.39)(mobile widget,0.42)(desktop graphic,0.05)};
\addplot[draw=pixcol!90,fill=pixcol!30] coordinates {(web text,0.59)(mobile widget,0.96)(desktop graphic,0.57)};
\addplot[draw=pixcol!90,fill=pixcol!12,
         postaction={pattern=north east lines,pattern color=pixcol!50}]
  coordinates {(web text,0.23)(mobile widget,0.98)(desktop graphic,0.60)};
\addplot[draw=envcol!90,fill=envcol!30] coordinates {(web text,0.21)(mobile widget,0.80)(desktop graphic,0.20)};
\addplot[draw=black!60,fill=black!12,
         postaction={pattern=dots,pattern color=black!50}]
  coordinates {(web text,0.49)(mobile widget,0.40)(desktop graphic,0.02)};
\legend{GPT-5.4,GPT-4o,Nano,Qwen-7B,Qwen3-30B,Intern-8B,OS-Atlas}
\end{axis}
\end{tikzpicture}
\caption{Pixel-following \PFR{} by family, all seven models. Non-text
mobile-widget identity stays pixel-bound for every capable general model;
only the smallest model and the grounding-tuned specialist collapse onto
structure there, ruling out a pure-OCR account.}
\label{fig:gradient_plot}
\end{figure}

\subsection{Non-text identity tracks capability and training}
If the substitution were string matching it would appear on every family,
but Figure~\ref{fig:gradient_plot} shows otherwise. On mobile widgets every
capable general-purpose model keeps following pixels, with $\PFR$ between
$0.80$ and $0.98$ across five models from three vendors. Two models break
the pattern, and each break is informative. The smallest model,
gpt-5.4-nano, resolves non-text conflicts toward structure, following it on
$0.58$ of mobile and $0.95$ of desktop probes while still recognizing the
icon at $\Accimg{=}0.94$; weaker visual capability therefore \emph{widens}
reliance on structure rather than degrading perception. The grounding-tuned
OS-Atlas-7B, fine-tuned to act on serialized GUI elements, likewise follows
structure on $0.60$ of mobile-widget and $0.98$ of desktop conflicts despite
seeing the icons: training a model to consume structure teaches it to trust
structure. Because desktop-graphic has the weakest gold validity, we
re-score it on the clean subset of $46$ probes both annotators confirm:
every model's \SFR{} moves in the same direction and the family ordering is
preserved, so the desktop conclusions are not artifacts of label noise.

\subsection{White-box ablation and what modulates the effect}
\label{sec:whitebox}
The behavioral gap shows \emph{that} beliefs defer to structure; we next ask
\emph{what} drives it, using Qwen2.5-VL-7B on the $230$ headline conflict
probes. A teacher-forced forced choice comparing the model's log-probability
of the two option strings reproduces its behavior, with a white-box \SFR{}
of $0.44$ against a behavioral $0.41$ on web text. We then zero the input
embeddings of exactly the tokens spelling the conflicting value inside the
structure block and re-decode, against zeroing an equal number of random
structure tokens. On web-text conflicts, removing the value flips the belief
back to the pixels in $0.63$ of structure-following cases versus $0.03$ for
random ablation, a roughly $19$-fold difference; a matched control that
zeroes an equal-length non-conflicting span flips only $0.02$, and the
effect replicates on OS-Atlas-Base-7B and InternVL3-8B at $0.56$ and $0.30$
versus $0.05$ and $0.08$ random. On desktop graphics the same removal flips
only $0.05$, near random, also on the clean desktop subset, so text
deference is driven by a single copied value while desktop deference is more
distributed. Prompt-level evidence points the same way: pruning the
structure or switching its serialization moves structure-following by at
most $0.03$, deleting the target value while keeping its context collapses
it from $0.55$ to $0.14$, and an explicit cue that the screenshot is the
authoritative source reverts $0.82$ of gpt-5.4's structure-following beliefs
and cuts its \PFG{} from $0.51$ to $0.05$, the closed-model analogue of the
embedding ablation. Our headline keeps a neutral prompt; the mitigation
comparison below shows why the prompt fix does not survive contact with
actions.

\subsection{From belief to action}
\label{sec:action}
We next show the mis-sourced belief changes what an agent \emph{does}, in
two live environments: AndroidWorld~\citep{rawles2024androidworld}, an
Android emulator running stock applications, and
MiniWoB++~\citep{liu2018miniwob}, a suite of scripted web tasks in a real
browser. The agent must click the element carrying a target label; we keep
the screenshot and all coordinates intact and swap only the target's
structural text with a distractor, so a pixel-sourced belief clicks the
visually correct element while a structure-sourced belief clicks the one
whose serialized text now reads the label. Under conflict the three open
models click the structure-named element on $0.80$ to $1.00$ of probes and
err on $0.91$ to $1.00$, against an aligned-channel error of at most $0.05$,
and no probe improves under conflict. Executing the chosen action confirms
real failure: the AndroidWorld tap lands on a wrong page on $25$ of $36$
audited probes, and the MiniWoB++ reward flips from ${\approx}{+}0.85$ when
aligned to negative under conflict. Within a single gpt-5.4 task of $120$
conflicts we split probes by the model's own stated belief; both groups face
the same swapped list, yet structure-belief probes are clicked wrong $0.65$
of the time versus $0.26$ for pixel-belief probes, an odds ratio of $5.34$
at $p{\approx}4\mathrm{e}{-}5$, so the wrong action follows the belief, not
the list corruption alone.

\subsection{One wrong belief compounds over a trajectory}
\label{sec:multistep}
Single actions understate the deployed risk, because an agent that mis-reads
one screen still faces every later screen with correct observations and
could in principle recover. We test this on live seeded MiniWoB++ episodes
of up to six clicks, drawn from twelve click-style task types and
prescreened to the $62$ seeded episodes per model that admit a valid
first-step conflict injection. Each seed runs under three paired conditions:
channels always aligned, swapped at every step, and, the diagnostic case,
swapped only at the first step with all later observations truthful.
Aligned, the three open models succeed on $0.77$ to $0.86$ of episodes.
Under a first-step-only conflict, episodes whose aligned twin needs a single
step recover often, at $0.41$ to $0.63$, since the next state reveals the
error, but episodes whose aligned twin needs two or more steps almost never
do: given a structure-following first step, final task failure is $0.97$ to
$1.00$ and recovery at most $0.03$ across Qwen2.5-VL-7B, InternVL3-8B, and
Qwen3-VL-30B. One mis-sourced belief is thus not a one-step tax but a
trajectory-level failure: the wrong click changes the page, and later
truthful observations do not pull the agent back.

\begin{table}[t]
\centering\small
\setlength{\tabcolsep}{3.5pt}
\begin{tabular}{@{}llccc@{}}
\toprule
 & & agree. & conflict & conflict \\
Model (acts via) & Env & pix. & str.-foll. & act.\ err \\
\midrule
\multirow{2}{*}{UGround-7B (coord)} & Web & 0.94 & \textbf{0.03} & \textbf{0.05} \\
 & Mobile & 0.77 & \textbf{0.07} & 0.36 \\
\multirow{2}{*}{Aguvis-7B (coord)} & Web & 0.89 & \textbf{0.04} & \textbf{0.11} \\
 & Mobile & 0.80 & \textbf{0.02} & \textbf{0.20} \\
\multirow{2}{*}{OS-Atlas-7B (index)} & Web & 0.99 & 0.97 & 1.00 \\
 & Mobile & 0.97 & 0.96 & 0.97 \\
\midrule
general VLMs (index) & Web & 0.99 & 0.95--1.00 & 0.96--1.00 \\
general VLMs (index) & Mobile & 0.95 & 0.80--0.93 & 0.91--0.98 \\
\bottomrule
\end{tabular}
\caption{Specialized GUI agents on the same conflict probes; Web is
MiniWoB++ and Mobile is AndroidWorld. The coordinate-emitting UGround and
Aguvis stay visually grounded under conflict, while the structure-indexed
OS-Atlas is hi\kern0.05pt jacked like the general VLMs: the split tracks
the action channel, not the model class.}
\label{tab:specialists}
\end{table}

\subsection{Specialized GUI agents}
\label{sec:specialists}
We delimit the scope of the effect along the action interface. Three open
specialists, all Qwen2-VL-7B based, UGround-V1-7B~\citep{gou2025uground},
Aguvis-7B~\citep{xu2025aguvis}, and OS-Atlas-Base-7B~\citep{wu2025osatlas},
split cleanly along \emph{how the agent emits its action} in
Table~\ref{tab:specialists}. UGround and Aguvis answer with a screen
coordinate and stay visually grounded under conflict, following structure on
$0.02$ to $0.07$ of probes, an order of magnitude below the general models;
OS-Atlas, which acts by naming an element in the serialized list, is
hijacked just as hard at $0.96$ to $0.97$. Holding Qwen2.5-VL-7B fixed and
swapping only its action-output format, structure-following rises
monotonically with how directly the action names the structured channel,
from $0.11$ for a coordinate to $0.64$ for an element id, $0.90$ for an
index, and $0.98$ for a text label, while the coordinate format's own
no-conflict error is the highest at $0.43$. Nor is coordinate action a
refuge: against an injected node with no rendered counterpart, a threat that
lives only in the structure, the coordinate agent still acts on it on
$0.84$ of probes on Qwen2.5-VL and $1.00$ on OS-Atlas, clicking an element
it cannot see. Neither pure channel is safe; the failure follows whichever
channel the action is bound to, with full sweeps in Appendix~\ref{app:specialists}.

\begin{table}[t]
\centering\small
\setlength{\tabcolsep}{2.1pt}
\begin{tabular}{@{}lccccc@{}}
\toprule
 & \multicolumn{2}{c}{text swap} & \multicolumn{2}{c}{stale node} & cost \\
\cmidrule(lr){2-3}\cmidrule(lr){4-5}
Arm & err & hijack & err & hijack & quer. \\
\midrule
baseline        & 0.96--1.00 & 0.93--0.98 & 0.76--0.99 & 0.69--0.95 & 0 \\
pixel-priority  & 0.95--1.00 & 0.92--0.97 & 0.73--0.99 & 0.63--0.95 & 0 \\
certificate     & 0.96--1.00 & 0.22--0.38 & 0.79--0.99 & \textbf{0.00} & 1 \\
consist.\ gate  & \textbf{0.56--0.75} & \textbf{0.28--0.43} & \textbf{0.46--0.66} & 0.03--0.07 & 1.8--2.0 \\
\bottomrule
\end{tabular}
\caption{Four mitigations on identical MiniWoB++ action probes, ranges over
four open models. The pixel-priority prompt barely moves the action-level
hijack; the certificate eliminates stale-node hijack but blocks
$0.59$--$0.77$ of conflicted actions; only the consistency gate reduces
hijack and task error together.}
\label{tab:mitigation}
\end{table}

\subsection{Comparing mitigations on identical probes}
\label{sec:gate}
The diagnosis suggests remedies at three levels, and we compare them on the
same action probes, models, and conditions rather than in isolation. A
coordinate-swap control that swaps element centers while keeping text
truthful first localizes the cause: baseline action error stays near zero,
so the hijack is driven by the structure's \emph{text}. The four arms are
the undefended baseline; a \emph{pixel-priority prompt} injected into the
action instruction; a \emph{certificate} check in the spirit of action-level
defenses~\citep{eca2026}, which crops the proposed element, asks whether the
pixels show the goal, and refuses the action when they do not; and our
\emph{source-aware consistency gate}, which performs the same cropped check
but on failure re-grounds by scanning on-screen elements for the one whose
pixels show the goal, so a stale node can never pass.
Table~\ref{tab:mitigation} reports the outcome. The prompt cue, so
effective at the belief level, leaves action-level hijack at $0.92$ to
$0.97$: the indexed action forces the model back onto the corrupted
channel regardless of the stated priority. The certificate eliminates stale-node
hijack but refuses $0.59$ to $0.77$ of conflicted actions, so its task
error stays at baseline: it converts hijacks into safe refusals but does
not complete tasks. The gate matches the certificate's hijack reduction
while also completing tasks, cutting text-swap error by $24$ to $44$
points and leaving aligned behavior within $0.04$ of baseline; on dense
AndroidWorld screens the repair is partial at $3.3$ to $5.5$ queries per
action, and the gate trusts geometry, so the coordinate-swap threat is out
of scope. No arm dominates; the comparison charts a cost-benefit frontier.

\subsection{Controls and robustness}
Further controls rule out shallow accounts. A reverse counter-family of
$180$ probes inverts the roles: the structure stays truthful and the target
region is blurred unreadable. Blur alone drops accuracy to chance, $0.54$
and $0.59$, and a correct structure recovers the true value at $0.95$ and
$1.00$, so models rely on structure exactly when pixels fail; the
over-reliance we measure under \emph{clear} pixels is therefore
mis-calibrated fusion, not an artifact of defining gold by the pixels. The
re-render control is answered at $1.00$, structure-following answers never
self-report pixel evidence on the frontier model, and \SFR{} varies by at
most $0.04$ over five temperature-$0.7$ seeds. Finally,
two human annotators given the same screenshot and structure on a
stratified $120$-probe subset read the on-screen value at $0.94$ to $0.95$
and follow the conflicting structure only $0.03$ of the time, far below
every model; humans err by not seeing, so structure-following is a model
failure, not task ambiguity.

%% file: sections/07_discussion.tex
\section{Discussion and Limitations}
\label{sec:discussion}

\paragraph{A diagnostic with measured consequences.}
Visual state reliance is a property of the \emph{belief}, so we frame it
as a diagnostic rather than a defense, but its consequences are measured,
not conjectured. The failure is two-sided:
structure-reading agents inherit whatever the channel asserts, while
coordinate action, which resists this hijack, acts on structure-only
threats it cannot see; the agent must know which channel its belief came
from. Capability does not settle it
either: scaling from 7B to 30B repaired visibility grounding on ghost
nodes yet worsened text-value deference, so the next model generation will
not simply outgrow the failure.

\paragraph{Limitations.}
Four caveats bound the claims. Closed-model corroboration spans one
vendor, and the capability contrast rests on open checkpoints from 7B to
30B, so the capability-reliance relation is charted, not a law. The compounding result uses click-style MiniWoB++ episodes of up to six
steps, though the mechanism, committing to an unrecoverable branch, is not
click-specific. The harshest
natural kind has $42$ consensus-audited stale web probes from $11$ sites,
so its hijack estimates carry moderate intervals; the mining pipeline can
grow the pool cheaply. Finally, the expanded families use model annotators under the same
two-annotator pixel-only protocol; the decisive mobile-stale subset and
the human baseline rest on human annotation.

%% file: sections/08_conclusion.tex
\section{Conclusion}
\label{sec:conclusion}

We introduced \task{}, attributing a state belief to pixels, structure, or
priors, with a $735$-probe benchmark of synthetic and natural conflicts
and the Perception-Fusion Gap. Across seven models,
textual beliefs defer to a conflicting structural value despite
near-ceiling image-only accuracy; the deference survives natural severity
and model scale, compounds one wrong step into trajectory failure, and no
compared mitigation resolves it without cost. Future agents should track
not just the predicted state but the channel that supports it.

%% file: sections/09_appendix.tex
This appendix carries the complete experimental protocols and results
behind the main text. Appendix~\ref{app:conditions} details the
interventions, Appendix~\ref{app:data} the construction pipeline including
the live-site mining and the severity distributions,
Appendix~\ref{app:metrics} the full per-model and per-platform metrics that
expand the main results table, Appendix~\ref{app:cluster} the cluster-robust
confidence intervals, Appendix~\ref{app:mcnemar} the complete cross-model
significance matrix, Appendix~\ref{app:ablation} the full prompt ablation by
family, Appendix~\ref{app:natural} the complete natural-conflict results,
Appendix~\ref{app:strict} the strict perception variant,
Appendix~\ref{app:specialists} the specialized agents and the action-format
sweep, Appendix~\ref{app:audit} the two-annotator gold-label audit and
label-noise sensitivity, Appendix~\ref{app:counter} the reverse
counter-family, Appendix~\ref{app:human} the human baseline,
Appendix~\ref{app:whitebox} the white-box attribution details,
Appendix~\ref{app:action} the belief-to-action, multi-step, and mitigation
details, and Appendix~\ref{app:prompt} the prompt and reproducibility
settings.

\section{Intervention and Scoring Details}
\label{app:conditions}
Each probe is rendered under the eight conditions of
Table~\ref{tab:conditions} of the main text. The headline structure-swap conflict keeps the real
screenshot and replaces the structural value with a distractor, so a model
that answers the distractor is following structure. The natural-conflict
condition is the zero-edit variant in which the conflict already exists in the
data, used for the mobile families. The pixel-swap condition perturbs the pixels
toward the distractor while keeping the true structural value, and its re-render
control re-renders the true value onto the same region to measure whether the
rendering operation alone changes the answer. The pixels-only condition exposes
the image only and measures perception. The structure-only condition exposes the
structure only and measures whether the model reads the structure. The
no-evidence condition occludes the region and removes the structure to expose the
prior. Options are shuffled per probe with a fixed seed, and answers are matched
to $a^\star$ or $b$ by string identity with no model judge.

\section{Dataset Construction Details}
\label{app:data}
\paragraph{Web text, Multimodal-Mind2Web.}
We scan task screenshots for a text node whose bounding box is verified
against the rendered page, take the on-element text as the pixel gold
$a^\star$, and form the distractor $b$ from another on-page string of the
same broad type. The screenshot is never edited for the conflict contrast.

\paragraph{Mobile widgets, stale references, and controls, RICO and CLAY.}
We align RICO view-hierarchy nodes with CLAY human annotations. For the
mobile-stale family we select nodes that CLAY marks invalid or background,
which are stale or ghost references in the tree. For the mobile-widget family we
select non-text widgets whose CLAY type label gives the pixel gold, such as icon,
image, or switch, and assert a text label in the structure. The visible positive
control selects nodes that CLAY marks valid and visible.

\paragraph{Desktop graphics, ScreenSpot-Pro.}
We crop the high-resolution screenshot around the target element, draw the
target region, take the human \texttt{ui\_type} label as the pixel gold,
namely icon or text label, and assert the opposite type in the structure.

\paragraph{Web-natural, live-site mining.}
A Selenium crawler visits $38$ live production websites, renders each page in
a fixed-size Chrome-for-Testing browser, and captures the serialized DOM and
a screenshot of the same viewport. It flags three kinds of zero-edit
divergence between the two channels. \emph{aria}: a visible element whose
\texttt{aria-label} shares no containment relation with its rendered text
after normalization. \emph{title}: the same test on the \texttt{title}
attribute. \emph{stale}: the DOM is serialized, the page is allowed to update
for a fixed delay, and the screenshot is taken afterward; elements whose
serialized text now differs from the rendered text, tickers, clocks, and
counters, are stale snapshots of the queried field. Candidates are filtered
for visibility, minimum text length, and on-viewport position, and each probe
archives the page URL, the full DOM, and the screenshot. The pool contains
$225$ probes, $55$ aria, $105$ title, and $65$ stale; scoring uses the
audit-consensus subset of $182$.

\paragraph{Mobile-stale expansion.}
The original release carried $50$ CLAY-flagged mobile-stale candidates.
We expand the pool to $250$ candidates over $99$ distinct RICO screens with
at most three probes per screen, and subject all $250$ to the human
visibility audit described in Appendix~\ref{app:audit}.

\paragraph{Counts.}
The benchmark has $735$ probes: $120$ web-text, $50$ mobile-widget, $250$
mobile-stale candidates, $30$ mobile visible controls, $60$ desktop-graphic,
and $225$ web-natural. The edited-conflict headline uses the $230$ probes of
the web-text, mobile-widget, and desktop-graphic families.

\paragraph{Severity distributions.}
For every conflict pair, the pixel value against the structural value, we
compute five string severity measures: normalized Levenshtein distance,
character-level difference ratio, token Jaccard distance, trigram cosine
distance, and relative length difference. Table~\ref{tab:appseverity}
compares the $120$ synthetic web-text pairs against the $225$ natural
web pairs. On every measure the synthetic median sits at or above the natural
median, meaning our edits are, if anything, \emph{easier} to distinguish from
the rendered value than natural divergence is, and $93$ to $100\%$ of
synthetic pairs fall inside the natural $5$th-to-$95$th percentile band.
Kolmogorov--Smirnov tests reject distributional identity, the synthetic
distribution is narrower, but the deference we measure on synthetic conflicts
cannot be attributed to implausibly harsh distractors.

\begin{table}[H]
\centering\small
\setlength{\tabcolsep}{5pt}
\begin{tabular}{@{}lcccc@{}}
\toprule
Measure & syn.\ median & nat.\ median & coverage & KS $p$ \\
\midrule
norm.\ edit dist.   & 0.89 & 0.80 & 1.00 & ${<}0.001$ \\
char.\ difference   & 0.81 & 0.66 & 1.00 & ${<}0.001$ \\
token Jaccard       & 1.00 & 1.00 & 0.99 & 0.021 \\
trigram cosine      & 1.00 & 0.85 & 1.00 & ${<}0.001$ \\
length difference   & 0.48 & 0.38 & 0.93 & ${<}0.001$ \\
\bottomrule
\end{tabular}
\caption{Severity of synthetic web-text distractors versus natural web
divergences ($120$ versus $225$ pairs). Coverage is the share of synthetic
pairs inside the natural $5$th-to-$95$th percentile band.}
\label{tab:appseverity}
\end{table}

\begin{figure}[H]
\centering
\includegraphics[width=\columnwidth]{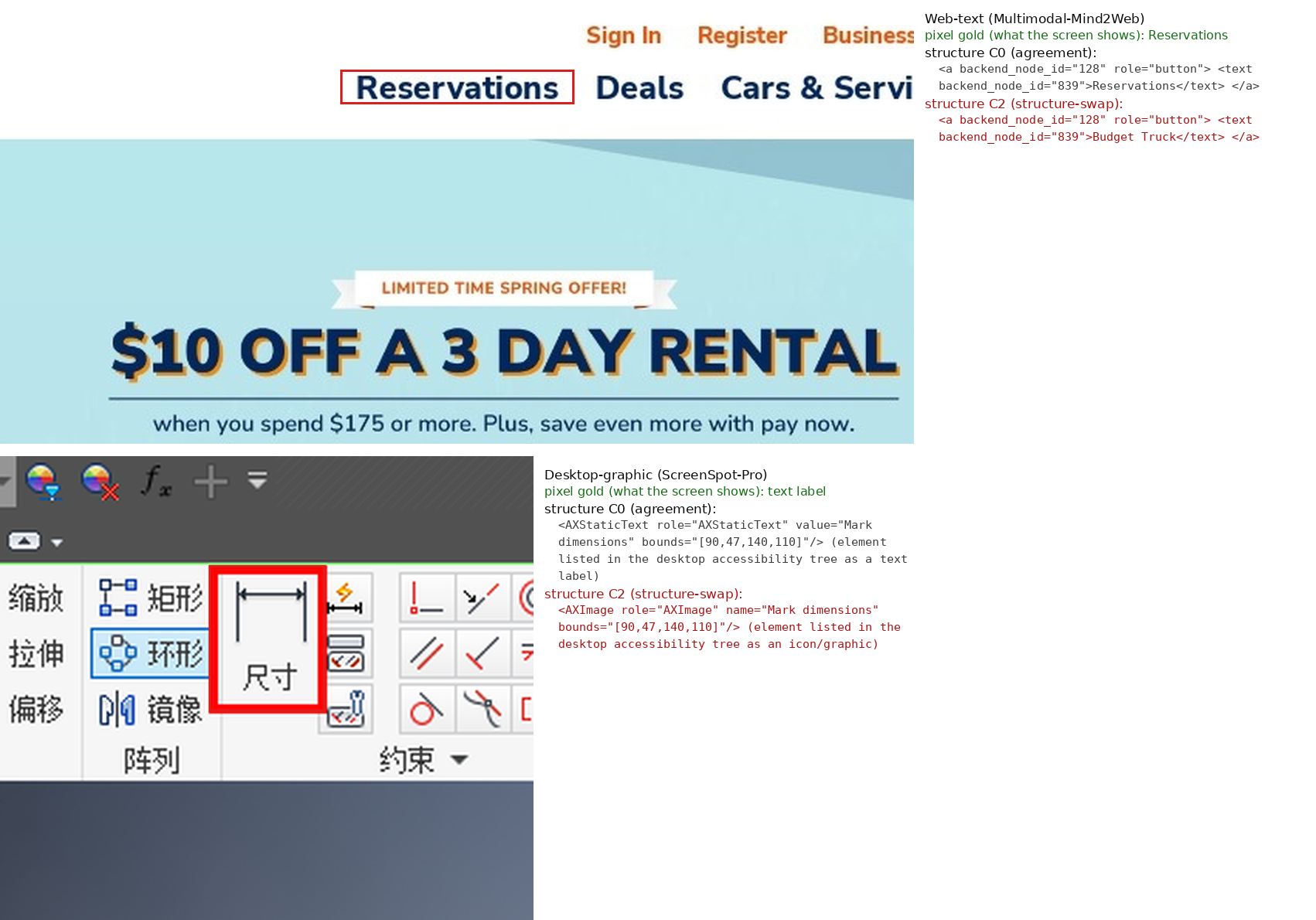}
\caption{Two real structure-swap probes. \textbf{Top:} a web-text probe from
Multimodal-Mind2Web; the screenshot shows \emph{Reservations} (outlined, the
pixel gold), while the C2 structure asserts the distractor \emph{Budget Truck}
with the screenshot untouched. \textbf{Bottom:} a desktop-graphic probe from
ScreenSpot-Pro; the toolbar element renders a text label (pixel gold), while the
C2 structure asserts the opposite \texttt{ui\_type} (an icon). In both cases only
the serialized structure is edited, so answering the C2 value is
structure-following. Images and structure strings are taken verbatim from the
benchmark.}
\label{fig:probeexamples}
\end{figure}

\section{Full Per-Model and Per-Platform Metrics}
\label{app:metrics}
Table~\ref{tab:appmetrics} reports every metric for the five models of the
initial evaluation wave, plus the structure-only reader, on the original
$310$-probe core, overall and per platform. The two open-weight models added
in the second wave, Qwen3-VL-30B and OS-Atlas-7B, were run on the full
benchmark with the identical protocol; their per-family results appear in the
main-text tables, in Table~\ref{tab:gradient}, and in the cluster and
significance appendices below, with complete per-probe records in the
released run manifests. Table~\ref{tab:appci} reports bootstrap $95\%$
confidence intervals on the overall set for the initial wave.

We first define the four supporting diagnostics summarized in the
Metrics section. \VOR{} (visual override resistance) is the complement of
\PFG{} on the perception-correct subset $\mathcal{C}$,
$\frac{1}{|\mathcal{C}|}\sum_{i\in\mathcal{C}}\mathbb{1}[\hat{y}^{\text{swap}}_i
{=}a^\star_i]$. \PPD{} (pixel-perturbation detection) is the rate of answering
the distractor under the pixel-swap condition,
$\frac1N\sum_i\mathbb{1}[\hat{y}^{\text{pxsw}}_i{=}b_i]$; low \PPD{} corroborates
high \SFR. \NVR{} compares \PFR{} on non-text versus text families. \SRM{}
(self-report mismatch) is the share of structure-following answers for which the
model self-reported \texttt{evidence\_channel}=\texttt{pixel}, probing
metacognitive consistency.

\begin{table*}[t]
\centering\small
\setlength{\tabcolsep}{5pt}
\begin{tabular}{@{}llccccccccc@{}}
\toprule
Model & Split & $\Accimg$ & $\Acczero$ & \PFR & \SFR & \PFG & \VOR & \PPD & re-render & str.-only \\
\midrule
\multirow{3}{*}{gpt-5.4}
 & overall & 0.877 & 0.954 & 0.636 & 0.364 & 0.288 & 0.712 & 0.350 & 1.000 & 0.968 \\
 & web     & 0.925 & 0.992 & 0.450 & 0.550 & 0.514 & 0.486 & 0.350 & 1.000 & 0.925 \\
 & mobile  & 0.808 & 0.900 & 0.830 & 0.170 & 0.000 & 1.000 & --    & --    & 1.000 \\
\midrule
\multirow{3}{*}{gpt-4o}
 & overall & 0.856 & 0.930 & 0.564 & 0.436 & 0.350 & 0.650 & 0.350 & 1.000 & 0.971 \\
 & web     & 0.916 & 1.000 & 0.283 & 0.717 & 0.688 & 0.312 & 0.350 & 1.000 & 0.933 \\
 & mobile  & 0.819 & 0.923 & 0.780 & 0.220 & 0.073 & 0.927 & --    & --    & 1.000 \\
\midrule
\multirow{3}{*}{structure-only}
 & overall & --    & 0.992 & 0.025 & 0.975 & --    & --    & --    & --    & 0.971 \\
 & web     & --    & 0.992 & 0.058 & 0.942 & --    & --    & --    & --    & 0.933 \\
 & mobile  & --    & 0.988 & 0.000 & 1.000 & --    & --    & --    & --    & 1.000 \\
\midrule
\multirow{3}{*}{gpt-5.4-nano}
 & overall & 0.842 & 0.977 & 0.290 & 0.710 & 0.662 & 0.333 & 0.358 & 1.000 & 0.946 \\
 & web     & 0.900 & 0.983 & 0.387 & 0.613 & 0.574 & 0.417 & 0.358 & 1.000 & 0.900 \\
 & mobile  & 0.785 & 0.963 & 0.320 & 0.680 & 0.603 & 0.397 & --    & --    & 0.970 \\
\midrule
\multirow{3}{*}{Qwen2.5-VL-7B}
 & overall & 0.816 & 0.969 & 0.547 & 0.453 & 0.327 & 0.664 & 0.633 & 0.992 & 0.954 \\
 & web     & 0.860 & 0.983 & 0.595 & 0.405 & 0.296 & 0.684 & 0.633 & 0.992 & 0.892 \\
 & mobile  & 0.762 & 1.000 & 0.480 & 0.520 & 0.351 & 0.649 & --    & --    & 1.000 \\
\midrule
\multirow{3}{*}{InternVL3-8B}
 & overall & 0.827 & 0.973 & 0.278 & 0.722 & 0.665 & 0.330 & 0.133 & 1.000 & 0.953 \\
 & web     & 0.892 & 0.983 & 0.214 & 0.786 & 0.747 & 0.242 & 0.133 & 1.000 & 0.891 \\
 & mobile  & 0.792 & 0.975 & 0.400 & 0.600 & 0.519 & 0.481 & --    & --    & 1.000 \\
\bottomrule
\end{tabular}
\caption{Full metrics by model and platform over the original $N{=}310$ probe
core and the structure-only reader. The web rows expand
Table~\ref{tab:main} of the main text. Dashes mark quantities undefined for a split, such as
image metrics for the structure-only reader or pixel perturbation on the mobile
visibility families.}
\label{tab:appmetrics}
\end{table*}

\begin{table}[H]
\centering\small
\setlength{\tabcolsep}{4pt}
\begin{tabular}{@{}lcccc@{}}
\toprule
Model & \PFR & \SFR & \PFG & \VOR \\
\midrule
gpt-5.4        & [.58,.69] & [.31,.42] & [.29,.40] & [.68,.78] \\
gpt-4o         & [.51,.62] & [.38,.49] & [.35,.46] & [.62,.73] \\
gpt-5.4-nano   & [.24,.34] & [.66,.76] & [.64,.75] & [.34,.45] \\
Qwen2.5-VL-7B  & [.49,.61] & [.39,.51] & [.34,.46] & [.64,.75] \\
InternVL3-8B   & [.23,.33] & [.67,.77] & [.63,.75] & [.33,.45] \\
structure-only & [.01,.05] & [.96,.99] & --        & --        \\
\bottomrule
\end{tabular}
\caption{Bootstrap $95\%$ confidence intervals on the overall set from
by-probe resampling.}
\label{tab:appci}
\end{table}

\begin{table}[H]
\centering\small
\setlength{\tabcolsep}{5pt}
\begin{tabular}{@{}llccc@{}}
\toprule
 & & web & mobile & desktop \\
Model & Vendor & text & widget & graphic \\
\midrule
gpt-5.4       & OpenAI    & 0.45 & 0.96 & 0.68 \\
gpt-4o        & OpenAI    & 0.28 & 0.98 & 0.77 \\
gpt-5.4-nano  & OpenAI    & 0.39 & 0.42 & 0.05 \\
Qwen2.5-VL-7B & Alibaba   & 0.59 & 0.96 & 0.57 \\
Qwen3-VL-30B  & Alibaba   & 0.23 & 0.98 & 0.60 \\
InternVL3-8B  & OpenGVLab & 0.21 & 0.80 & 0.20 \\
OS-Atlas-7B   & SH AI Lab & 0.49 & 0.40 & 0.02 \\
\bottomrule
\end{tabular}
\caption{Pixel-following \PFR{} by family under conflict on the three
headline families, across four vendors; these values underlie Figure~\ref{fig:gradient_plot} of the
main text. Higher means more reliance on pixels. Non-text mobile-widget
identity stays pixel-bound for every capable general model regardless of
vendor (\PFR{} $0.80$ to $0.98$); the two exceptions are the smallest model
(gpt-5.4-nano) and the grounding-tuned specialist OS-Atlas-7B, which the main
text discusses. Web text draws every model toward structure to differing
degrees. This cross-vendor pattern is what rules out a pure-OCR or
string-matching account of the text effect.}
\label{tab:gradient}
\end{table}

\section{Cluster-robust confidence intervals}
\label{app:cluster}
By-probe resampling can understate variance when probes share a website or
screen. We therefore recompute intervals for the headline web-text family with a
\emph{cluster} bootstrap ($5000$ resamples) that treats each Mind2Web
website/task hash as one cluster ($29$ clusters over $120$ probes) and resamples
clusters with replacement, for all seven models. Table~\ref{tab:appcluster}
reports the point estimate and cluster $95\%$ interval for \SFR{} and \PFG.
Point estimates are recomputed under the cluster protocol, which conditions
perception on the annotated gold value, and can therefore differ from the
main-text table by at most $0.02$. The \PFG{} interval excludes zero for every
model, so the perception--fusion gap survives the more conservative
clustering.

\begin{table}[H]
\centering\small
\setlength{\tabcolsep}{4pt}
\begin{tabular}{@{}lcc@{}}
\toprule
Model & \SFR{} (pt [lo, hi]) & \PFG{} (pt [lo, hi]) \\
\midrule
Qwen2.5-VL-7B & 0.41 [0.29, 0.53] & 0.30 [0.17, 0.45] \\
Qwen3-VL-30B  & 0.77 [0.66, 0.87] & 0.73 [0.60, 0.84] \\
InternVL3-8B  & 0.79 [0.72, 0.85] & 0.76 [0.68, 0.82] \\
OS-Atlas-7B   & 0.51 [0.42, 0.61] & 0.48 [0.38, 0.57] \\
gpt-5.4       & 0.55 [0.46, 0.64] & 0.51 [0.42, 0.61] \\
gpt-4o        & 0.72 [0.64, 0.79] & 0.69 [0.60, 0.77] \\
gpt-5.4-nano  & 0.61 [0.50, 0.72] & 0.58 [0.47, 0.69] \\
\bottomrule
\end{tabular}
\caption{Cluster-robust $95\%$ intervals for the web-text family, resampling by
website/task cluster ($29$ clusters, $120$ probes, $5000$ bootstrap draws). The
\PFG{} interval is strictly positive for all seven models.}
\label{tab:appcluster}
\end{table}

\section{Cross-Model Significance Matrix}
\label{app:mcnemar}
Table~\ref{tab:appmcnemar} gives the paired McNemar tests on web
structure-following, where $b$ counts probes the first model of the pair
follows structure and the second does not and $c$ counts the reverse; Holm
correction is applied jointly over all ten tests. Both frontier models and all
four open-weight models fall below the structure-only ceiling at
$p_{\text{Holm}}<0.001$, which is the test referenced in the main bracketing
result. Pair sizes below $120$ reflect probes on which one model returned no
parsable answer.

\begin{table}[H]
\centering\small
\setlength{\tabcolsep}{4pt}
\begin{tabular}{@{}lccccc@{}}
\toprule
Pair & $n$ & $b$ & $c$ & $p_{\text{raw}}$ & $p_{\text{Holm}}$ \\
\midrule
gpt-5.4 vs gpt-4o          & 120 & 8  & 28 & 0.0012 & 0.0036 \\
gpt-5.4 vs structure-only  & 120 & 1  & 48 & ${<}0.001$ & ${<}0.001$ \\
gpt-5.4 vs gpt-5.4-nano    & 119 & 13 & 20 & 0.296  & 0.296 \\
gpt-4o vs structure-only   & 120 & 1  & 28 & ${<}0.001$ & ${<}0.001$ \\
gpt-4o vs gpt-5.4-nano     & 119 & 24 & 11 & 0.041  & 0.082 \\
structure-only vs nano     & 119 & 40 & 1  & ${<}0.001$ & ${<}0.001$ \\
Qwen2.5-VL-7B vs structure-only & 116 & 0 & 63 & ${<}0.001$ & ${<}0.001$ \\
Qwen3-VL-30B vs structure-only  & 115 & 1 & 21 & ${<}0.001$ & ${<}0.001$ \\
InternVL3-8B vs structure-only  & 117 & 2 & 21 & ${<}0.001$ & ${<}0.001$ \\
OS-Atlas-7B vs structure-only   & 113 & 1 & 50 & ${<}0.001$ & ${<}0.001$ \\
\bottomrule
\end{tabular}
\caption{Exact McNemar tests with Holm correction on web-text
structure-following.}
\label{tab:appmcnemar}
\end{table}

\section{Full Ablation by Family}
\label{app:ablation}
Table~\ref{tab:appablation} reports structure-following under conflict for
gpt-5.4 across all four families and all six prompt variants; the body cites the
web-text, desktop-graphic, and overall columns. Pixel-following is one minus the
listed value on these conflict conditions.

\begin{table}[H]
\centering\small
\setlength{\tabcolsep}{4pt}
\begin{tabular}{@{}lccccc@{}}
\toprule
Ablation & web & mobile & mob.-stale & desktop & all \\
\midrule
baseline          & 0.55 & 0.04 & 0.30 & 0.32 & 0.36 \\
dom\_pruned       & 0.58 & 0.08 & 0.34 & 0.30 & 0.39 \\
dom\_ax\_only     & 0.54 & 0.04 & 0.32 & 0.28 & 0.36 \\
image\_first      & 0.24 & 0.02 & 0.30 & 0.23 & 0.21 \\
stale\_label      & 0.08 & 0.02 & 0.34 & 0.12 & 0.13 \\
source\_of\_truth & 0.10 & 0.06 & 0.28 & 0.12 & 0.13 \\
context\_only     & 0.14 & 0.02 & 0.32 & 0.07 & 0.14 \\
dom\_only         & 0.94 & 1.00 & 1.00 & 1.00 & 0.98 \\
\bottomrule
\end{tabular}
\caption{Structure-following \SFR{} under conflict for gpt-5.4 by family and
prompt variant over $280$ conflict probes (the mobile-stale family is shown
for completeness only). Serialization variants stay near baseline, while channel
ordering, the staleness and source-of-truth cues, and value removal lower the
gap. The \texttt{dom\_only} row withholds the screenshot and marks the
structure-following upper bound. The mobile-widget family is already near zero at
baseline, so the frontier model rarely defers on mobile icons regardless of the
variant.}
\label{tab:appablation}
\end{table}

\section{Complete Natural-Conflict Results}
\label{app:natural}
Table~\ref{tab:appnatural} expands the natural-conflict table of the main
text with sample sizes, pixel-following rates, and Wilson $95\%$ lower
bounds, on the $182$-probe audit-consensus subset of the web-natural family
($48$ aria, $92$ title, $42$ stale; per-cell $n$ counts scored answers).
Table~\ref{tab:appmobilestale} reports the mobile-stale family on the
$114$-probe subset both human auditors unanimously confirm absent from the
screen. \emph{Absence detection} is image-only accuracy at reporting the
ghost element not visible; the last column conditions, as in \PFG{}, on the
probes whose image-only read already detects the absence and reports how
often the fused belief still flips to \emph{visible}.

\begin{table}[H]
\centering\small
\setlength{\tabcolsep}{3pt}
\begin{tabular}{@{}lcccccc@{}}
\toprule
 & \multicolumn{2}{c}{aria} & \multicolumn{2}{c}{title} & \multicolumn{2}{c}{stale} \\
\cmidrule(lr){2-3}\cmidrule(lr){4-5}\cmidrule(lr){6-7}
Model & \SFR & $n$ & \SFR & $n$ & \SFR{} [lo] & $n$ \\
\midrule
Qwen2.5-VL-7B & 0.02 & 44 & 0.03 & 92 & 0.38 [0.25] & 42 \\
Qwen3-VL-30B  & 0.02 & 47 & 0.01 & 92 & 0.76 [0.62] & 42 \\
InternVL3-8B  & 0.05 & 43 & 0.02 & 91 & 0.88 [0.75] & 42 \\
OS-Atlas-7B   & 0.02 & 46 & 0.08 & 91 & 0.45 [0.31] & 42 \\
\bottomrule
\end{tabular}
\caption{Web-natural \SFR{} by divergence kind on the audit-consensus
subset, with Wilson $95\%$ lower bounds for the stale kind. The $42$ stale
probes come from $11$ distinct sites. Pixel-following is one minus \SFR{}
up to unparsed answers; image-only accuracy is $0.94$ to $0.96$ for all
four models.}
\label{tab:appnatural}
\end{table}

\begin{table}[H]
\centering\small
\setlength{\tabcolsep}{4pt}
\begin{tabular}{@{}lcccc@{}}
\toprule
Model & \SFR{} [lo] & abs.\ det. & flip$\mid$perc.\ [lo] & $n_{\text{perc}}$ \\
\midrule
Qwen2.5-VL-7B & 0.99 [0.95] & 0.45 & 0.98 [0.90] & 51 \\
Qwen3-VL-30B  & 0.19 [0.13] & 0.75 & 0.02 [0.01] & 85 \\
InternVL3-8B  & 0.94 [0.88] & 0.77 & 0.93 [0.86] & 88 \\
OS-Atlas-7B   & 0.75 [0.66] & 0.60 & 0.63 [0.51] & 68 \\
\bottomrule
\end{tabular}
\caption{Mobile-stale family on the $114$ unanimously-confirmed-absent
probes. \SFR{} is the rate of reporting the ghost element visible under
fused inputs; ``abs.\ det.''\ is image-only absence detection;
``flip$\mid$perc.''\ conditions on correct image-only detection. Wilson
$95\%$ lower bounds in brackets. Qwen3-VL-30B is largely immune here while
being the most hijacked model on synthetic web text, so capability repairs
visibility grounding but not text-value fusion.}
\label{tab:appmobilestale}
\end{table}

\section{Strict Perception Variant}
\label{app:strict}
Table~\ref{tab:appstrict} reports the crop-verified strict variant of the
gap for all four open models, on the web-text family and over the full
benchmark. $\Acccrop$ is accuracy answering from the tight crop of the
target region alone; $\PFG_{\text{strict}}$ recomputes the gap on the
crop-verified subset. The strict variant stays within $0.05$ of \PFG{}
everywhere and rises on web text for three of the four models, so the gap
is not an artifact of full-screenshot perception failures. The gradient
attribution check runs on the three locally instrumented models: among
web-text conflicts whose belief follows structure, the mean
gradient$\times$input attribution mass per visual token is non-zero on
every probe and comparable to pixel-following cases, $0.028$ versus
$0.034$ on Qwen2.5-VL-7B, $0.018$ versus $0.016$ on InternVL3-8B, and
$0.024$ versus $0.023$ on OS-Atlas-7B, so the visual evidence is processed
and then overridden at fusion rather than never read.

\begin{table}[H]
\centering\small
\setlength{\tabcolsep}{3.5pt}
\begin{tabular}{@{}lccccc@{}}
\toprule
 & \multicolumn{3}{c}{web text} & \multicolumn{2}{c}{overall} \\
\cmidrule(lr){2-4}\cmidrule(lr){5-6}
Model & $\Acccrop$ & \PFG & $\PFG_{\text{str}}$ & \PFG & $\PFG_{\text{str}}$ \\
\midrule
Qwen2.5-VL-7B & 0.93 & 0.30 & 0.34 & 0.33 & 0.33 \\
Qwen3-VL-30B  & 0.92 & 0.71 & 0.77 & 0.24 & 0.26 \\
InternVL3-8B  & 0.91 & 0.75 & 0.75 & 0.58 & 0.57 \\
OS-Atlas-7B   & 0.93 & 0.45 & 0.44 & 0.46 & 0.47 \\
\bottomrule
\end{tabular}
\caption{Strict crop-verified perception check. $\Acccrop$ exceeds the
full-screenshot $\Accimg$ on web text for every model, and
$\PFG_{\text{strict}}$ tracks \PFG{} within $0.05$, so the gap survives
the stricter perception gate.}
\label{tab:appstrict}
\end{table}

\section{Specialists and the Action-Format Sweep}
\label{app:specialists}
The specialist table of the main text is complete as printed; this
appendix adds the two supporting sweeps on AndroidWorld, $256$ probes per
format. Table~\ref{tab:appformat} holds the model fixed and varies only
the action-output format. Structure-following under conflict rises
monotonically with how directly the action names the structured channel,
while the coordinate format pays the highest no-conflict error.
Table~\ref{tab:appinjected} shows the injected-node threat: a structural
entry with no rendered counterpart is added to the element list and the
agent is asked to act on its label. Every format, including the
coordinate one, acts on the phantom element at $0.84$ or above, so
coordinate action is not a refuge against threats that live only in the
structure.

\begin{table}[H]
\centering\small
\setlength{\tabcolsep}{3.5pt}
\begin{tabular}{@{}lcccccc@{}}
\toprule
 & \multicolumn{3}{c}{Qwen2.5-VL-7B} & \multicolumn{3}{c}{OS-Atlas-7B} \\
\cmidrule(lr){2-4}\cmidrule(lr){5-7}
Format & C0 err & SF & C2 err & C0 err & SF & C2 err \\
\midrule
coordinate & 0.43 & 0.11 & 0.50 & 0.77 & 0.12 & 0.93 \\
element-id & 0.12 & 0.64 & 0.84 & 0.12 & 0.66 & 0.88 \\
index      & 0.04 & 0.90 & 0.95 & 0.04 & 0.95 & 0.97 \\
text-label & 0.02 & 0.98 & 1.00 & 0.02 & 0.98 & 1.00 \\
\bottomrule
\end{tabular}
\caption{Action-format sweep on AndroidWorld, $256$ probes per format,
same model and screens. ``SF'' is structure-following under the text-swap
conflict; ``C0 err'' the aligned-channel action error. Structure-following
rises monotonically with the directness of the structural binding, and the
coordinate format's own no-conflict error is the highest.}
\label{tab:appformat}
\end{table}

\begin{table}[H]
\centering\small
\setlength{\tabcolsep}{6pt}
\begin{tabular}{@{}lcc@{}}
\toprule
Format & Qwen2.5-VL-7B & OS-Atlas-7B \\
\midrule
coordinate & 0.84 & 1.00 \\
element-id & 0.94 & 0.96 \\
index      & 0.95 & 1.00 \\
text-label & 1.00 & 1.00 \\
\bottomrule
\end{tabular}
\caption{Rate of acting on an injected structural node with no rendered
counterpart, by action format ($256$ probes per cell). The agent clicks an
element it cannot see under every format.}
\label{tab:appinjected}
\end{table}

\section{Gold-Label Audit}
\label{app:audit}
Two independent annotators audited every probe against the rendered
pixels, seeing only the screenshot and never the structure, and choosing the
pixel value from the option set or marking the probe \emph{unsure} or
\emph{bad}. Validity is the share of audited probes whose pixel judgment
matches the proposed gold, excluding \emph{unsure} and \emph{bad}. Over all
$735$ probes the two annotators reach $0.880$ raw agreement and Cohen's
$\kappa{=}0.86$. Table~\ref{tab:appaudit} reports per-family agreement and
validity. Validity is $0.94$ on web text, $0.96$ on mobile non-text widgets,
$0.96$ to $0.97$ on web-natural, and $0.82$ to $0.88$ on desktop document
graphics, so the headline and natural families are well supported.
The desktop-graphic family carries the lowest headline validity
($0.82$--$0.88$); inspecting its disagreements, every one is an
icon-versus-text-label boundary call on toolbar buttons that render both an icon
and a nearby text label, not arbitrary annotator noise, so the residual
uncertainty is confined to a single well-defined category rather than spread
across the family. The mobile-stale family is handled differently by design:
its proposed gold asserts the \emph{absence} of a CLAY-flagged node, CLAY
flags are noisy, and pixel annotators frequently find the node still
rendered. Two human auditors therefore annotated all $250$ candidates for
visibility, agreeing at $0.912$ raw agreement and $\kappa{=}0.825$, and only
the $114$ elements both auditors confirm absent enter any score. The
web-natural family is scored on the analogous audit-consensus subset, $182$
of $225$, because stale ticker values are small on screen and the strict
consensus rule removes any probe whose gold either annotator could not read
back from the pixels.

\begin{table}[H]
\centering\small
\setlength{\tabcolsep}{6pt}
\begin{tabular}{@{}lcccc@{}}
\toprule
Family & $n$ & agreement & $\kappa$ & validity (A/B) \\
\midrule
web-text              & 120 & 0.84 & 0.84 & 0.94 / 0.94 \\
mobile-widget         & 50  & 1.00 & 1.00 & 0.96 / 0.96 \\
desktop-graphic       & 60  & 0.83 & 0.67 & 0.88 / 0.82 \\
visible control       & 30  & 0.83 & --   & 1.00 / 1.00 \\
mobile-stale (cand.)  & 250 & 0.88 & --   & audit-gated \\
web-natural           & 225 & 0.89 & 0.89 & 0.97 / 0.96 \\
\bottomrule
\end{tabular}
\caption{Two-annotator gold-label audit over all $735$ probes. Annotators see
only pixels. Validity is each annotator's agreement with the proposed gold,
excluding probes marked unsure or bad. The $\kappa$ dash marks families where
one rating is degenerate or the gold is an absence claim. Mobile-stale is
scored only on the $114$-probe unanimously-confirmed-absent subset from the
dedicated human visibility audit; web-natural is scored on its $182$-probe
audit-consensus subset.}
\label{tab:appaudit}
\end{table}

\paragraph{Label-noise sensitivity.}
To show the headline results do not depend on residual label noise, we re-run
them on the \emph{gold-confirmed} subset, the $202$ probes whose proposed gold
both annotators independently confirm. The white-box value-ablation effect is
preserved: on the gold-confirmed subset, removing the conflicting value flips the
belief in $0.38$ of structure-following cases against $0.06$ for random ablation,
versus $0.33$ against $0.04$ on the full set. The behavioral structure-following
and the belief-to-action propagation likewise hold on this subset, so the
conclusions are robust to the $\sim 6$ to $19\%$ label noise across families.

\paragraph{Desktop clean subset.}
Because desktop-graphic carries the weakest validity, the main text re-scores
it on the clean subset of $46$ probes both annotators confirm.
Table~\ref{tab:appdeskclean} gives the full per-model comparison. Every
model's \SFR{} moves in the same direction from the full family to the clean
subset and the cross-model ordering is preserved, so the desktop conclusions,
including the near-random white-box value flip on this family, are not
artifacts of label noise.

\begin{table}[H]
\centering\small
\setlength{\tabcolsep}{6pt}
\begin{tabular}{@{}lcc@{}}
\toprule
Model & \SFR{} full ($n{=}60$) & \SFR{} clean ($n{=}46$) \\
\midrule
gpt-5.4        & 0.32 & 0.13 \\
gpt-4o         & 0.23 & 0.04 \\
gpt-5.4-nano   & 0.95 & 0.94 \\
Qwen2.5-VL-7B  & 0.43 & 0.35 \\
InternVL3-8B   & 0.80 & 0.74 \\
OS-Atlas-7B    & 0.98 & 0.98 \\
\bottomrule
\end{tabular}
\caption{Desktop-graphic \SFR{} under conflict on the full family versus the
clean subset both annotators confirm. The ordering is unchanged.}
\label{tab:appdeskclean}
\end{table}

\section{Reverse Counter-Family}
\label{app:counter}
Table~\ref{tab:counter} reports the reverse counter-family of the
main text (structure correct, pixels blurred unreadable, $N{=}180$).

\begin{table}[H]
\centering\small
\setlength{\tabcolsep}{4pt}
\begin{tabular}{@{}lcccc@{}}
\toprule
 & blur-only & true-struct & false-struct & conflict \\
Model & acc & recovery & parrot & \SFR \\
\midrule
Qwen2.5-VL-7B & 0.54 & 0.95 & 0.84 & 0.45 \\
InternVL3-8B  & 0.59 & 1.00 & 0.92 & 0.72 \\
\bottomrule
\end{tabular}
\caption{Reverse counter-family (structure correct, pixels blurred unreadable),
$N{=}180$. \emph{Blur-only acc} near chance shows the blur removes the pixel
signal; \emph{true-struct rec.}\ (recovery) near one shows the model correctly
switches to the truthful structure when pixels fail (credited by the pixel
gold); \emph{false-struct parrot} is the rate of repeating a wrong structural
value when pixels are unreadable. The high \emph{conflict \SFR} (clear pixels,
wrong structure) therefore reflects over-trust only when pixels are informative,
not a scoring artifact.}
\label{tab:counter}
\end{table}

\section{Human Baseline on the Conflict Task}
\label{app:human}
Distinct from the pixel-only gold audit above, we run a human baseline on the
\emph{full} conflict task: annotators see the same screenshot \emph{and} the
serialized structure and pick the on-screen value under the two-option forced
choice, with an explicit ``cannot tell'' abstention. We draw a stratified subset
of $120$ probes ($40$ each from the web-text, mobile-widget, and desktop-graphic
headline families) and collect two independent expert annotators.
Table~\ref{tab:apphuman} reports per-platform accuracy (chose the on-screen pixel
value), human \SFR{} (chose the conflicting structural value), abstention rate,
and inter-annotator agreement. Aggregated, human accuracy is $0.94$--$0.95$,
human \SFR{} is $0.033$, and agreement is Cohen's $\kappa{=}0.97$. Human
ambiguity concentrates entirely on small low-resolution web crops; on mobile and
desktop accuracy is $0.98$--$1.00$ with no abstentions and per-family agreement
$0.975$--$1.00$. The few non-pixel answers are dominated by abstentions and by
near-synonymous or subjective option pairs such as ``Profile'' versus ``Profile
menu'' or icon versus text label; excluding these, the genuine human
structure-following rate is $\approx 0.008$. Humans thus err by \emph{not seeing}
and abstain on unreadable crops, the opposite of the models, which err by reading
a legible screen and copying the structure. The annotation instrument, a
self-contained browser tool with target-region highlighting and keyboard entry,
and the scorer are released with the code.

\begin{table}[H]
\centering\small
\setlength{\tabcolsep}{5pt}
\resizebox{\columnwidth}{!}{%
\begin{tabular}{@{}lccccc@{}}
\toprule
Platform & $n$ & accuracy (A/B) & human \SFR{} (A/B) & abstain (A/B) & agreement \\
\midrule
web-text        & 40  & 0.90 / 0.85 & 0.05 / 0.05  & 0.05 / 0.10 & 0.95 \\
mobile-widget   & 40  & 0.98 / 1.00 & 0.03 / 0.00  & 0.00 / 0.00 & 0.98 \\
desktop-graphic & 40  & 0.98 / 0.98 & 0.03 / 0.03  & 0.00 / 0.00 & 1.00 \\
\midrule
all             & 120 & 0.95 / 0.94 & 0.03 / 0.03  & 0.02 / 0.03 & $\kappa{=}0.97$ \\
\bottomrule
\end{tabular}}
\caption{Human baseline on the full conflict task (two expert annotators A/B,
stratified $120$-probe subset). Annotators see the screenshot \emph{and} the
structure and pick the on-screen value. Humans overwhelmingly read the pixels
(human \SFR{} $0.03$), an order of magnitude below every model ($0.30$--$0.79$);
the residual is dominated by honest abstentions on unreadable web crops.}
\label{tab:apphuman}
\end{table}

\section{White-Box Attribution Details}
\label{app:whitebox}
We run Qwen2.5-VL-7B-Instruct locally on the $230$ headline conflict probes
(web-text, mobile-widget, desktop-graphic) under the structure-swap conflict. We
form the white-box forced choice by teacher-forcing the option strings and
comparing $\log p(a^\star\mid\text{prompt})$ and $\log p(b\mid\text{prompt})$,
taking the larger as the model's belief; this reproduces structure-following with
a white-box \SFR{} of $0.42$ overall and $0.44$ on web text. For the causal test
we delimit the serialized structure with sentinel markers, locate the token span
of the conflicting value $b$ inside it, zero those input embeddings, and
re-decode the forced choice; the control zeroes an equal number of randomly
chosen structure tokens. Table~\ref{tab:appwb} reports, per family, the
white-box \SFR{}, the number of structure-following probes in which the value
span is located, and the flip-to-pixel rate under value ablation versus random
ablation. Web-text conflicts show a large targeted-over-random gap, desktop
graphics do not, and the mobile-widget control family has too few
structure-following cases to read.

\begin{table}[H]
\centering\small
\setlength{\tabcolsep}{5pt}
\begin{tabular}{@{}lcccc@{}}
\toprule
Family & wb \SFR & located & value flip & random flip \\
\midrule
web-text        & 0.44 & 41/53 & 0.63 & 0.03 \\
desktop-graphic & 0.62 & 37/37 & 0.05 & 0.04 \\
mobile-widget   & 0.14 & 7/7   & 0.00 & 0.00 \\
\bottomrule
\end{tabular}
\caption{White-box causal ablation on Qwen2.5-VL-7B under the structure-swap
conflict. ``located'' is
the count of structure-following probes whose value span is found in the
serialized structure. ``value flip'' zeroes that span; ``random flip'' zeroes
an equal number of random structure tokens. Text conflicts have a localizable
causal driver; document graphics do not.}
\label{tab:appwb}
\end{table}

\paragraph{A black-box prompt-level analogue on the closed model.}
Embedding ablation is impossible on the closed OpenAI endpoints, so we cannot
run the white-box test on gpt-5.4. We instead apply the \emph{same
source-authority logic at the prompt level} and measure the paired belief flip.
Prepending an explicit cue that the screenshot is the authoritative source
(\texttt{source\_of\_truth}) is the pixel-priority system prompt a defensive
scaffold would add; restricted to the baseline structure-following probes, it
reverts the belief to the rendered value in $0.82$ of web-text and $0.63$ of
desktop-graphic cases (Table~\ref{tab:appblackbox}), the same direction as the
open-model value ablation ($0.63$ on web text) and larger. Correspondingly, the
perception-correct \PFG{} falls from $0.51$ to $0.05$ on web text and from
$0.27$ to $0.05$ on desktop. The value-redaction analogue
(\texttt{context\_only}, aggregate) lowers web-text \SFR{} from $0.55$ to $0.14$.
The closed-model evidence is thus behavioral/interventional at the prompt level
rather than a white-box embedding edit, but it corroborates the same localizable,
correctable cause; the residual \PFG{} is non-zero, and the cue must be explicitly
supplied and is not assumed by our neutral scaffold, so the neutral-prompt
headline reflects that scaffold rather than a worst case.

\begin{table}[H]
\centering\small
\setlength{\tabcolsep}{4pt}
\begin{tabular}{@{}lcccc@{}}
\toprule
Family & $n_{\mathrm{SF}}$ & flip$\to$pixel & \PFG{} base & \PFG{} sot \\
\midrule
web-text        & 66 & 0.82 & 0.51 & 0.05 \\
desktop-graphic & 19 & 0.63 & 0.27 & 0.05 \\
\bottomrule
\end{tabular}
\caption{Black-box prompt-level causal probe on gpt-5.4 (no new API calls; $230$
headline conflict probes). $n_{\mathrm{SF}}$ is the baseline
structure-following count; \emph{flip$\to$pixel} is the share of those probes
that revert to the pixel value under a pixel-priority (\texttt{source\_of\_truth})
prompt; \PFG{} is the perception-correct gap before and after the cue. The
mobile-widget family has too few baseline structure-following probes
($n_{\mathrm{SF}}{=}2$) to read. This is the closed-model analogue of the
white-box ablation in Table~\ref{tab:appwb}.}
\label{tab:appblackbox}
\end{table}

\paragraph{Why we do not report a saliency ratio.}
We also computed a structure-over-pixel attribution ratio, the structure share
of summed and per-token gradient$\times$input and squared-gradient (Fisher)
attribution over the image and structure token spans. This quantity is
unreliable here: image tokens, which are continuous post-vision-tower features,
and text tokens, which are embedding lookups, live on different scales, so
per-token Fisher mass concentrates on structure tokens almost regardless of
behavior, with a mean ratio near $0.99$, and the gradient$\times$input ratio
correlates with structure-following at $|r|\lesssim 0.05$. We therefore exclude the ratio from
the evidence and rely on the scale-free causal ablation above.

\section{Belief-to-Action Details}
\label{app:action}
We run belief-to-action in two live agent environments:
AndroidWorld~\citep{rawles2024androidworld}, a KVM-accelerated Android emulator
with stock apps, and MiniWoB++~\citep{liu2018miniwob}, a set of interactive web
tasks in a headless Chrome. For each captured screen we record the rendered
screenshot as
the pixel channel and the on-screen element list with text and center coordinate
as the structure channel, drawn from the Android accessibility tree or the web
DOM. A probe asks
the agent to \emph{tap the element labeled $X$}; in the structure-swap conflict we
swap the structural \emph{text} of the target element and a distractor while
leaving the screenshot and all coordinates untouched, so only the semantic label
disagrees. A pixel-sourced belief taps the element visually showing $X$
(\texttt{gold\_index}); a structure-sourced belief taps the element whose
serialized text now reads $X$ (\texttt{struct\_index}). The action is
\texttt{click(index)} into the element list, parsed deterministically; decoding is
greedy. We build $256$ AndroidWorld probes from $13$ screens (stock apps and
Settings subpages) and $470$ MiniWoB++ probes from $10$ task types, $726$ probes
in total; $22\%$ of the web probes also carry real hidden/stale DOM nodes.
Table~\ref{tab:appaction} reports the full per-environment, per-model numbers;
all McNemar tables are fully directional ($c{=}0$).

For the live closed loop we execute the chosen tap. On AndroidWorld we replay the
tap on the emulator and compare the resulting page title to that of the correct
tap; on $38$ audited probes the model follows structure on all $38$, and on the
$36$ with a verifiable landing page it reaches a different page on $25$ ($69\%$).
On MiniWoB++ each episode is started for real, the chosen
element is clicked in the browser, and the environment's reward is read back over
${\approx}90$ episodes per condition per model; aligned channels yield $0.95$ to
$0.97$ success (mean reward ${\approx}{+}0.85$) while the conflict yields $0.85$
to $1.00$ task failure (mean reward $-0.58$ to $-0.85$).

\begin{table}[H]
\centering\small
\setlength{\tabcolsep}{4pt}
\resizebox{\columnwidth}{!}{%
\begin{tabular}{@{}llcccc@{}}
\toprule
 & & agree. & conflict & conflict & McNemar \\
Env & Model & err & str.-foll. & act.\ err & $b/c$ \\
\midrule
\multirow{3}{*}{AndroidWorld} & Qwen2.5-VL-7B & 0.04 & 0.91 & 0.96 & 235/0 \\
 & InternVL3-8B & 0.05 & 0.93 & 0.98 & 239/0 \\
 & Qwen3-VL-30B & 0.05 & 0.80 & 0.91 & 220/0 \\
\midrule
\multirow{3}{*}{MiniWoB++} & Qwen2.5-VL-7B & 0.01 & 1.00 & 1.00 & 466/0 \\
 & InternVL3-8B & 0.01 & 0.95 & 0.96 & 449/0 \\
 & Qwen3-VL-30B & 0.00 & 0.98 & 1.00 & 467/0 \\
\bottomrule
\end{tabular}}
\caption{Belief-to-action under conflict across two live environments and three
open models. agree.\ err is the aligned-channel action-error rate; conflict
str.-foll.\ the fraction tapping the structure-named element; conflict act.\ err
the wrong-action rate under conflict. $b$ is the number of paired probes that go
correct$\to$wrong from the agreement to the conflict condition and $c$ the number
that go wrong$\to$correct.}
\label{tab:appaction}
\end{table}

\begin{table}[H]
\centering\small
\setlength{\tabcolsep}{3.5pt}
\resizebox{\columnwidth}{!}{%
\begin{tabular}{@{}lcccc@{}}
\toprule
 & \multicolumn{2}{c}{text-swap} & \multicolumn{2}{c}{stale-node} \\
\cmidrule(lr){2-3}\cmidrule(lr){4-5}
Model & err$\downarrow$ & SF$\downarrow$ & err$\downarrow$ & SF$\downarrow$ \\
\midrule
Qwen2.5-VL-7B & 1.00$\to$\textbf{0.56} & 0.98$\to$0.28 & 0.93$\to$\textbf{0.46} & 0.82$\to$0.06 \\
InternVL3-8B  & 0.96$\to$\textbf{0.60} & 0.93$\to$0.32 & 0.98$\to$\textbf{0.51} & 0.95$\to$0.07 \\
Qwen3-VL-30B  & 1.00$\to$\textbf{0.69} & 0.98$\to$0.38 & 0.76$\to$\textbf{0.50} & 0.69$\to$0.03 \\
OS-Atlas-7B   & 0.99$\to$\textbf{0.75} & 0.96$\to$0.43 & 0.99$\to$\textbf{0.66} & 0.93$\to$0.06 \\
\bottomrule
\end{tabular}}
\caption{Consistency-gate repair on the web (baseline$\to$gated). ``err'' is the
text-swap / stale-node action error, ``SF'' is structure-following. The gate
removes most of the hi\kern0.05pt jack (SF collapses) and lowers action error substantially;
the agreement-condition error is unchanged in every case ($\le 0.04$, omitted).
AndroidWorld shows the same direction but a smaller effect (consistency-gate section).}
\label{tab:gate}
\end{table}

\paragraph{Multi-step episodes.}
The multi-step experiment of the main text runs live seeded MiniWoB++
episodes under three paired conditions per seed: channels aligned (C0), the
structural texts of target and distractor swapped at every step, and swapped
only at the first step with all later observations truthful. Twelve
click-style task types are prescreened per seed, keeping episodes where the
first screen admits a valid swap, which yields $62$ paired episodes per model
across four surviving task types: click-button-sequence ($30$), navigate-tree
($17$), click-tab-2 ($14$), and click-widget ($1$).
Table~\ref{tab:appmultistep} reports per-model aligned success, success under
each conflict schedule, and the recovery decomposition. Episodes whose
aligned twin needs one step recover at $0.41$ to $0.63$; episodes needing two
or more steps fail at $0.97$ to $1.00$ given a structure-following first
step, with recovery at most $0.03$.

\begin{table}[H]
\centering\small
\setlength{\tabcolsep}{3.5pt}
\resizebox{\columnwidth}{!}{%
\begin{tabular}{@{}lcccccc@{}}
\toprule
 & C0 & first-step & all-step & s1 str.- & fail$\mid$str. & recov.$\mid$str. \\
Model & succ. & succ. & succ. & foll. & (multi) & (multi / single) \\
\midrule
Qwen2.5-VL-7B & 0.84 & 0.29 & 0.00 & 0.88 & 1.00 & 0.00 / 0.63 \\
InternVL3-8B  & 0.77 & 0.29 & 0.08 & 0.71 & 0.97 & 0.03 / 0.43 \\
Qwen3-VL-30B  & 0.86 & 0.21 & 0.11 & 0.89 & 1.00 & 0.00 / 0.41 \\
\bottomrule
\end{tabular}}
\caption{Multi-step MiniWoB++ episodes, $62$ paired seeds per model.
``s1 str.-foll.'' is the rate of structure-following first steps under the
first-step-only conflict; ``fail$\mid$str.'' and ``recov.$\mid$str.'' condition
on such a first step, split by whether the aligned twin needs two or more
steps (multi) or one (single).}
\label{tab:appmultistep}
\end{table}

\paragraph{Mitigation comparison, per model and environment.}
Table~\ref{tab:appmitig} expands the mitigation table of the main text to all
four open models and both environments. The qualitative ordering is identical
everywhere: the pixel-priority prompt leaves action-level hijack essentially
at baseline, the certificate eliminates stale-node hijack by refusal, and the
consistency gate is alone in lowering hijack and task error together, with a
weaker and costlier repair on dense AndroidWorld screens.

\begin{table}[H]
\centering\small
\setlength{\tabcolsep}{3pt}
\resizebox{\columnwidth}{!}{%
\begin{tabular}{@{}llcccccc@{}}
\toprule
 & & \multicolumn{3}{c}{MiniWoB++} & \multicolumn{3}{c}{AndroidWorld} \\
\cmidrule(lr){3-5}\cmidrule(lr){6-8}
Model & Arm & C2 err & hijack & blocked & C2 err & hijack & blocked \\
\midrule
\multirow{4}{*}{Qwen2.5-VL-7B}
 & baseline & 1.00 & 0.98 & 0.00 & 0.94 & 0.89 & 0.00 \\
 & prompt   & 0.95 & 0.94 & 0.00 & 0.95 & 0.89 & 0.00 \\
 & cert.    & 1.00 & 0.22 & 0.77 & 0.97 & 0.14 & 0.80 \\
 & gate     & 0.56 & 0.28 & 0.00 & 0.81 & 0.77 & 0.00 \\
\midrule
\multirow{4}{*}{Qwen3-VL-30B}
 & baseline & 1.00 & 0.98 & 0.00 & 0.97 & 0.66 & 0.00 \\
 & prompt   & 0.99 & 0.97 & 0.00 & 0.97 & 0.62 & 0.00 \\
 & cert.    & 1.00 & 0.33 & 0.65 & 0.98 & 0.16 & 0.54 \\
 & gate     & 0.69 & 0.38 & 0.00 & 0.94 & 0.61 & 0.00 \\
\midrule
\multirow{4}{*}{InternVL3-8B}
 & baseline & 0.96 & 0.93 & 0.00 & 0.98 & 0.93 & 0.00 \\
 & prompt   & 0.95 & 0.92 & 0.00 & 0.98 & 0.92 & 0.00 \\
 & cert.    & 0.96 & 0.25 & 0.69 & 0.99 & 0.29 & 0.68 \\
 & gate     & 0.60 & 0.32 & 0.00 & 0.84 & 0.75 & 0.00 \\
\midrule
\multirow{4}{*}{OS-Atlas-7B}
 & baseline & 0.99 & 0.96 & 0.00 & 0.96 & 0.95 & 0.00 \\
 & prompt   & 1.00 & 0.97 & 0.00 & 0.97 & 0.95 & 0.00 \\
 & cert.    & 0.99 & 0.38 & 0.59 & 0.96 & 0.36 & 0.60 \\
 & gate     & 0.75 & 0.43 & 0.00 & 0.68 & 0.55 & 0.00 \\
\bottomrule
\end{tabular}}
\caption{Per-model mitigation comparison under the text-swap conflict (C2).
``hijack'' is the rate of acting on the structure-named element, ``blocked''
the refusal rate. Gate extra queries per action are $1.8$ to $2.0$ on
MiniWoB++ and $3.3$ to $5.5$ on AndroidWorld; certificate always costs one.
Aligned-condition (C0) error for prompt and gate stays within $0.04$ of
baseline on MiniWoB++, while the certificate raises C0 error to $0.06$ to
$0.10$ there and up to $0.48$ on AndroidWorld through false refusals.}
\label{tab:appmitig}
\end{table}

\paragraph{Action error tracks the belief, not the corrupted list.}
Within a single gpt-5.4 click task ($120$ web-text conflicts) we split probes by
the model's own stated belief; both groups see the identical swapped index list.
Table~\ref{tab:action_or} shows structure-belief probes err $2.5\times$ more
often, so the wrong action follows the belief rather than the list corruption.

\begin{table}[H]
\centering\small
\setlength{\tabcolsep}{6pt}
\begin{tabular}{@{}lccc@{}}
\toprule
Stated belief & $n$ & wrong click & not-found \\
\midrule
follows structure & 66 & 0.65 & 0.35 \\
follows pixels     & 54 & 0.26 & 0.74 \\
\bottomrule
\end{tabular}
\caption{Within one gpt-5.4 click task ($120$ web-text conflicts), wrong-click
rate split by the model's stated belief. Both groups see the identical swapped
index list, yet structure-belief probes err $2.5\times$ more often (odds ratio
$5.34$, $\chi^2_{\text{Yates}}{=}16.8$, $p{\approx}4\mathrm{e}{-}5$), so the
action error follows the belief and is not an artifact of the corrupted list.}
\label{tab:action_or}
\end{table}

\section{Prompt and Reproducibility}
\label{app:prompt}
The model receives a neutral instruction, the screenshot when the condition
exposes pixels, the serialized structure when the condition exposes
structure, and a forced-choice question with deterministically shuffled
options. It returns JSON with an \texttt{answer} in the value set, a
\texttt{confidence}, and a self-reported \texttt{evidence\_channel} in the set
pixel, structure, both, or unsure. Only the deterministic match of the answer
enters the metrics, and the self-report enters only the \SRM{} analysis. All
runs use temperature $0$ and a single sample per probe and condition. The four
open-weight models, Qwen2.5-VL-7B-Instruct, Qwen3-VL-30B-A3B-Instruct,
InternVL3-8B, and OS-Atlas-Base-7B, are run locally from fixed public
checkpoints (commit hashes released with the code) and constitute the
reproducible core; all four cover every probe of the full $735$-probe
benchmark, including the expanded natural families. The specialist agents
UGround-V1-7B and Aguvis-7B are likewise run locally from fixed checkpoints.
The three OpenAI models
(\texttt{gpt-5.4}, \texttt{gpt-5.4-nano}, \texttt{gpt-4o}) are accessed through
an OpenAI-compatible interface using the standard rolling model aliases; we did
not pin dated snapshots. All OpenAI calls were issued in a single window in June
2026 at temperature $0$ with one sample per probe and condition, and we log the
request timestamps and every returned \texttt{system\_fingerprint} in the
released run manifests. Because these aliases are served by rolling endpoints
that can drift, exact reproduction of the OpenAI numbers is not guaranteed; for
this reason the three OpenAI models are reported only as corroboration on the
original $310$-probe core, and every headline claim is established on the four
locally hosted open-weight models from fixed checkpoints. The structure-only reader is \texttt{gpt-5.4} with the
screenshot withheld, and no language model grades any answer or generates any
probe, label, or structure.

%% file: refs.bib
@inproceedings{deng2023mind2web,
  title={Mind2Web: Towards a Generalist Agent for the Web},
  author={Deng, Xiang and others},
  booktitle={NeurIPS},
  year={2023},
  note={arXiv:2306.06070}
}

@inproceedings{koh2024visualwebarena,
  title={VisualWebArena: Evaluating Multimodal Agents on Realistic Visual Web Tasks},
  author={Koh, Jing Yu and others},
  booktitle={ACL},
  year={2024},
  note={arXiv:2401.13649}
}

@inproceedings{xie2024osworld,
  title={OSWorld: Benchmarking Multimodal Agents for Open-Ended Tasks in Real Computer Environments},
  author={Xie, Tianbao and others},
  booktitle={NeurIPS},
  year={2024},
  note={arXiv:2404.07972}
}

@article{rawles2024androidworld,
  title={AndroidWorld: A Dynamic Benchmarking Environment for Autonomous Agents},
  author={Rawles, Christopher and others},
  journal={arXiv preprint arXiv:2405.14573},
  year={2024}
}

@inproceedings{liu2018miniwob,
  title={Reinforcement Learning on Web Interfaces Using Workflow-Guided Exploration},
  author={Liu, Evan Zheran and Guu, Kelvin and Pasupat, Panupong and Shi, Tianlin and Liang, Percy},
  booktitle={ICLR},
  year={2018},
  note={arXiv:1802.08802}
}

@article{chen2024guiworld,
  title={GUI-World: A Video Benchmark and Dataset for Multimodal GUI-Oriented Understanding},
  author={Chen, Dongping and others},
  journal={arXiv preprint arXiv:2406.10819},
  year={2024}
}

@inproceedings{cheng2024seeclick,
  title={SeeClick: Harnessing GUI Grounding for Advanced Visual GUI Agents},
  author={Cheng, Kanzhi and others},
  booktitle={ACL},
  year={2024},
  note={arXiv:2401.10935}
}

@article{li2025screenspotpro,
  title={ScreenSpot-Pro: GUI Grounding for Professional High-Resolution Computer Use},
  author={Li, Kaixin and others},
  journal={arXiv preprint arXiv:2504.07981},
  year={2025}
}

@article{d2snap2025,
  title={Beyond Pixels: Exploring DOM Downsampling for LLM-Based Web Agents},
  author={Schiepanski, Thassilo M. and Pi{\"e}l, Nicholas},
  journal={arXiv preprint arXiv:2508.04412},
  year={2025}
}

@article{screen2ax2025,
  title={Screen2AX: Vision-Based Approach for Automatic macOS Accessibility Generation},
  author={Muryn, Viktor and Sumyk, Marta and Hirna, Mariya and Garkot, Sofiya and Shamrai, Maksym},
  journal={arXiv preprint arXiv:2507.16704},
  year={2025}
}

@article{morethinking2025,
  title={More Thinking, Less Seeing? Assessing Amplified Hallucination in Multimodal Reasoning Models},
  author={Liu, Chengzhi and Xu, Zhongxing and Wei, Qingyue and Wu, Juncheng and Zou, James and Wang, Xin Eric and Zhou, Yuyin and Liu, Sheng},
  journal={arXiv preprint arXiv:2505.21523},
  year={2025}
}

@article{steplevel2026,
  title={Step-Level Visual Grounding Faithfulness Predicts Out-of-Distribution Generalization in Long-Horizon Vision-Language Models},
  author={Rahman, Md Ashikur and Rahman, Md Arifur and Samin, Niamul Hassan and Arean, Abdullah Ibne Hanif and Noshin, Juena Ahmed},
  journal={arXiv preprint arXiv:2603.06828},
  year={2026}
}

@article{drivingvla2026,
  title={Grounding Driving VLA via Inverse Kinematics},
  author={Park, Junsung and Shim, Hyunjung},
  journal={arXiv preprint arXiv:2605.21061},
  year={2026}
}

@article{memeye2026,
  title={MemEye: A Visual-Centric Evaluation Framework for Multimodal Agent Memory},
  author={Guo, Minghao and Jiao, Qingyue and Shi, Zeru and Quan, Yihao and Zhang, Boxuan and Li, Danrui and Che, Liwei and Xu, Wujiang and Liu, Shilong and Liu, Zirui and Kapadia, Mubbasir and Pavlovic, Vladimir and Liu, Jiang and Wang, Mengdi and Shi, Yiyu and Metaxas, Dimitris N. and Tang, Ruixiang},
  journal={arXiv preprint arXiv:2605.15128},
  year={2026}
}

@article{li2025vpibench,
  title={VPI-Bench: Visual Prompt Injection Attacks for Computer-Use Agents},
  author={Cao, Tri and Lim, Bennett and Liu, Yue and Sui, Yuan and Li, Yuexin and Deng, Shumin and Lu, Lin and Oo, Nay and Yan, Shuicheng and Hooi, Bryan},
  journal={arXiv preprint arXiv:2506.02456},
  year={2025}
}

@article{eva2025,
  title={EVA: Evolving Semantic Adversaries for Red-Teaming GUI Agents Against Environmental Injection Attacks},
  author={Lu, Yijie and Zhao, Manman and Ju, Tianjie and Yan, Zihe and Ma, Xinbei and Guo, Yuan and Ding, Daizong and Liu, Gongshen and Zhang, Zhuosheng},
  journal={arXiv preprint arXiv:2505.14289},
  year={2025}
}

@article{envinjection2025,
  title={WebInject: Prompt Injection Attack to Web Agents},
  author={Wang, Xilong and Bloch, John and Shao, Zedian and Hu, Yuepeng and Zhou, Shuyan and Gong, Neil Zhenqiang},
  journal={arXiv preprint arXiv:2505.11717},
  year={2025}
}

@article{osharm2025,
  title={OS-Harm: A Benchmark for Measuring Safety of Computer Use Agents},
  author={Kuntz, Thomas and Duzan, Agatha and Zhao, Hao and Croce, Francesco and Kolter, Zico and Flammarion, Nicolas and Andriushchenko, Maksym},
  journal={arXiv preprint arXiv:2506.14866},
  year={2025}
}

@inproceedings{deka2017rico,
  title={Rico: A Mobile App Dataset for Building Data-Driven Design Applications},
  author={Deka, Biplab and others},
  booktitle={UIST},
  year={2017}
}

@inproceedings{li2022clay,
  title={Learning to Denoise Raw Mobile UI Layouts for Improving Datasets at Scale},
  author={Li, Gang and others},
  booktitle={CHI},
  year={2022}
}

@article{eca2026,
  title={Hallucination as Exploit: Evidence-Carrying Multimodal Agents},
  author={Zhang, Guijia and Zheng, Hao and Yang, Harry},
  journal={arXiv preprint arXiv:2605.19192},
  year={2026}
}

@inproceedings{gou2025uground,
  title={Navigating the Digital World as Humans Do: Universal Visual Grounding for GUI Agents},
  author={Gou, Boyu and others},
  booktitle={ICLR},
  year={2025},
  note={arXiv:2410.05243}
}

@inproceedings{takeshita2026a11y,
  title={A11y-Compressor: A Framework for Enhancing the Efficiency of GUI Agent Observations through Visual Context Reconstruction and Redundancy Reduction},
  author={Takeshita, Michito and Kawada, Takuro and Ohashi, Takumi and Kitada, Shunsuke and Iyatomi, Hitoshi},
  booktitle={ACL (Student Research Workshop)},
  year={2026}
}

@inproceedings{zheng2024seeact,
  title={GPT-4V(ision) is a Generalist Web Agent, if Grounded},
  author={Zheng, Boyuan and Gou, Boyu and Kil, Jihyung and Sun, Huan and Su, Yu},
  booktitle={ICML},
  year={2024},
  note={arXiv:2401.01614}
}

@inproceedings{he2024webvoyager,
  title={WebVoyager: Building an End-to-End Web Agent with Large Multimodal Models},
  author={He, Hongliang and Yao, Wenlin and Ma, Kaixin and Yu, Wenhao and Dai, Yong and Zhang, Hongming and Lan, Zhenzhong and Yu, Dong},
  booktitle={ACL},
  year={2024},
  note={arXiv:2401.13919}
}

@article{qin2025uitars,
  title={UI-TARS: Pioneering Automated GUI Interaction with Native Agents},
  author={Qin, Yujia and others},
  journal={arXiv preprint arXiv:2501.12326},
  year={2025}
}

@inproceedings{xu2025aguvis,
  title={Aguvis: Unified Pure Vision Agents for Autonomous GUI Interaction},
  author={Xu, Yiheng and Wang, Zekun and Wang, Junli and Lu, Dunjie and Xie, Tianbao and Saha, Amrita and Sahoo, Doyen and Yu, Tao and Xiong, Caiming},
  booktitle={ICML},
  year={2025},
  note={arXiv:2412.04454}
}

@inproceedings{wu2025osatlas,
  title={OS-ATLAS: A Foundation Action Model for Generalist GUI Agents},
  author={Wu, Zhiyong and Wu, Zhenyu and Xu, Fangzhi and Wang, Yian and Sun, Qiushi and Jia, Chengyou and Cheng, Kanzhi and Ding, Zichen and Chen, Liheng and Liang, Paul Pu and Qiao, Yu},
  booktitle={ICLR},
  year={2025},
  note={arXiv:2410.23218}
}

@inproceedings{lin2025showui,
  title={ShowUI: One Vision-Language-Action Model for GUI Visual Agent},
  author={Lin, Kevin Qinghong and Li, Linjie and Gao, Difei and Yang, Zhengyuan and Wu, Shiwei and Bai, Zechen and Lei, Weixian and Wang, Lijuan and Shou, Mike Zheng},
  booktitle={CVPR},
  year={2025},
  note={arXiv:2411.17465}
}

@article{lu2024omniparser,
  title={OmniParser for Pure Vision Based GUI Agent},
  author={Lu, Yadong and Yang, Jianwei and Shen, Yelong and Awadallah, Ahmed},
  journal={arXiv preprint arXiv:2408.00203},
  year={2024}
}

@inproceedings{hong2024cogagent,
  title={CogAgent: A Visual Language Model for GUI Agents},
  author={Hong, Wenyi and Wang, Weihan and Lv, Qingsong and Xu, Jiazheng and Yu, Wenmeng and Ji, Junhui and Wang, Yan and Wang, Zihan and Dong, Yuxiao and Ding, Ming and Tang, Jie},
  booktitle={CVPR},
  year={2024},
  note={arXiv:2312.08914}
}

@article{wu2025guiactor,
  title={GUI-Actor: Coordinate-Free Visual Grounding for GUI Agents},
  author={Wu, Qianhui and Cheng, Kanzhi and Yang, Rui and Zhang, Chaoyun and Yang, Jianwei and Jiang, Huiqiang and Mu, Jian and Peng, Baolin and Qiao, Bo and Zheng, Liqun and others},
  journal={arXiv preprint arXiv:2506.03143},
  year={2025}
}

@article{yuan2025segui,
  title={Scaling Computer-Use Grounding via User Interface Decomposition and Synthesis},
  author={Yuan, Zhengyu and others},
  journal={arXiv preprint arXiv:2505.12370},
  year={2025}
}

@article{vcd2026,
  title={Visual Confused Deputy: Exploiting and Defending Perception Failures in Computer-Using Agents},
  author={Liu, X. and He, Bowei and Liu, Xue and Luo, Andy and Zhang, Haichen and Chen, Huamin},
  journal={arXiv preprint arXiv:2603.14707},
  year={2026}
}

@article{jia2025osworldmcp,
  title={OSWorld-MCP: Benchmarking MCP Tool Invocation in Computer-Use Agents},
  author={Jia, Hongrui and Liao, Jitong and Zhang, Xi and Xu, Haiyang and Xie, Tianbao and Jiang, Chaoya and Yan, Ming and Liu, Si and Ye, Wei and Huang, Fei},
  journal={arXiv preprint arXiv:2510.24563},
  year={2025}
}
